\documentclass[10pt,twocolumn,letterpaper]{article}

\usepackage{iccv}
\usepackage{times}
\usepackage{epsfig}
\usepackage{graphicx}
\usepackage{amsmath}
\usepackage{amssymb}
\usepackage{subfigure}
\usepackage{url}

\newcommand{\dsname}[1]{\texttt{\small #1}\xspace}

\newcommand{\scrfd}{\dsname{SCRFD}}
\newcommand{\scrfdv}[1]{\dsname{SCRFD\textsubscript{#1}}}
\newcommand{\scrfdvf}[2]{\dsname{SCRFD\textsubscript{#1}-\text{#2}GF}}
\newcommand{\scrfdf}[1]{\dsname{SCRFD-\text{#1}GF}}
\usepackage{enumitem}
\newenvironment{tight_itemize}{
\begin{itemize}[leftmargin=20pt]
  \setlength{\topsep}{0pt}
  \setlength{\itemsep}{0pt}
  \setlength{\parskip}{0pt}
  \setlength{\parsep}{0pt}
}{\end{itemize}}

\newcommand*{\affaddr}[1]{#1} 
\newcommand*{\affmark}[1][*]{\textsuperscript{#1}}
\renewcommand\thefootnote{}
\usepackage[pagebackref=true,breaklinks=true,letterpaper=true,colorlinks,bookmarks=false]{hyperref}

\iccvfinalcopy 

\def\iccvPaperID{0000} 

\ificcvfinal\fi

\begin{document}


\title{Sample and Computation Redistribution for Efficient Face Detection}

\author{
Jia Guo \textsuperscript{*} \affmark[2] \qquad 
Jiankang Deng \textsuperscript{*} \affmark[1,2] \qquad
Alexandros Lattas\affmark[1] \qquad Stefanos Zafeiriou\affmark[1]\\
\affaddr{\affmark[1]Imperial College} \qquad
\affaddr{\affmark[2]InsightFace} \qquad \\
{\tt\small guojia@gmail.com, \{j.deng16, a.lattas,s.zafeiriou\}@imperial.ac.uk}
}

\maketitle
\ificcvfinal\thispagestyle{empty}\fi

\begin{abstract}
Although tremendous strides have been made in uncontrolled face detection, efficient face detection with a low computation cost as well as high precision remains an open challenge. In this paper, we point out that training data sampling and computation distribution strategies are the keys to efficient and accurate face detection. Motivated by these observations, we introduce two simple but effective methods (1) Sample Redistribution (SR), which augments training samples for the most needed stages, based on the statistics of benchmark datasets; and (2) Computation Redistribution (CR), which reallocates the computation between the backbone, neck and head of the model, based on a meticulously defined search methodology. Extensive experiments conducted on WIDER FACE demonstrate the state-of-the-art efficiency-accuracy trade-off for the proposed \scrfd family across a wide range of compute regimes. In particular, \scrfdf{34} outperforms the best competitor, TinaFace, by $3.86\%$ (AP at hard set) while being more than \emph{3$\times$ faster} on GPUs with VGA-resolution images. We also release our code to facilitate future research. \url{https://github.com/deepinsight/insightface/tree/master/detection/scrfd}

\footnote{\textsuperscript{*} Equal contributions. 
InsightFace is a nonprofit face project.} 
\setcounter{footnote}{0}
\renewcommand\thefootnote{\arabic{footnote}}
\end{abstract}

\section{Introduction}
Face detection is a long-standing problem in computer vision with many applications, such as face alignment~\cite{bulat2017far}, face reconstruction \cite{feng2018joint}, face attribute analysis~\cite{zhang2018jointexpression,pan2018mean}, and face recognition~\cite{schroff2015facenet,deng2018arcface}.
Following the pioneering work of Viola-Jones \cite{viola2004robust}, numerous face detection algorithms have been designed. 
Among them, the single-shot anchor-based approaches~\cite{najibi2017ssh,zhang2017s3fd,tang2018pyramidbox,li2019dsfd,ming2019group,deng2019retinaface,liu2019hambox,zhu2020tinaface} have recently demonstrated the most promising performance. In particular, on the most challenging face detection dataset WIDER FACE \cite{yang2016wider}, the average precision (AP) on its Hard test set has been boosted to $92.4\%$ by TinaFace \cite{zhu2020tinaface}. 

\begin{figure}[t!]
\centering
\includegraphics[width=0.35\textwidth]{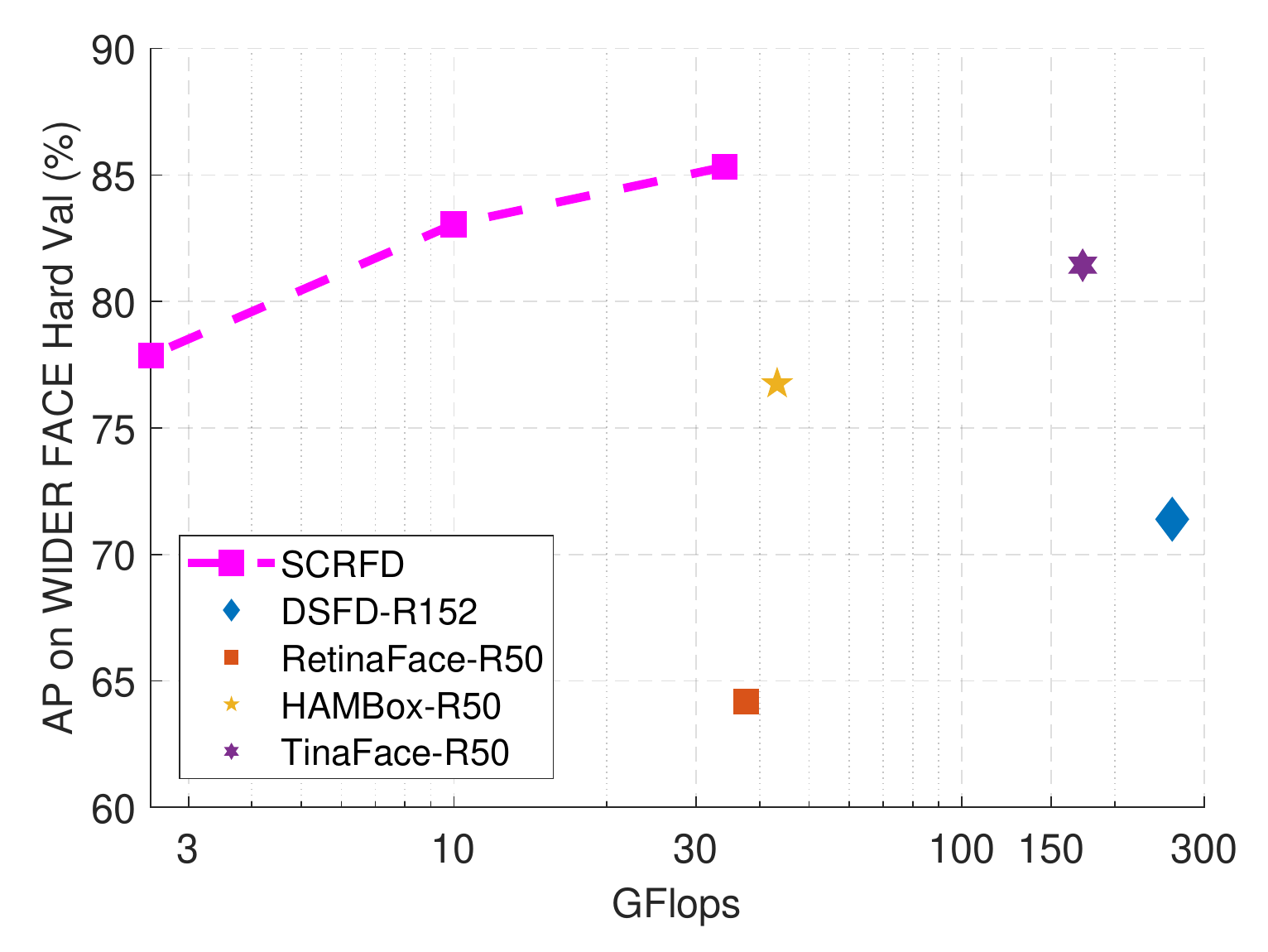}
\caption{Performance-computation trade-off on the WIDER FACE validation hard set for different face detectors. Flops and APs are reported by using the VGA resolution ($640\times480$) during testing. The proposed \scrfd outperforms a range of state-of-the-art open-sourced methods by using much fewer flops.}
\vspace{-4mm}
\label{fig:evelope}
\end{figure}

Even though TinaFace \cite{zhu2020tinaface} achieves impressive results on unconstrained face detection, it employs large-scale (\eg $1,650$ pixels) testing, which consumes huge amounts of computational cost (as shown in Tab.~ \ref{tab:tinafacedifscale}). In addition, TinaFace is designed based on a generic object detector (\ie RetinaNet \cite{lin2017focal}), directly taking the classification network as the backbone, tiling dense anchors on the multi-scale feature maps (\ie P2 to P7 of neck) and adopting heavy head designs. Without considering the prior of faces, the backbone, neck and head design of TinaFace is thus redundant and sub-optimal. 

Since directly taking the backbone of the classification network for object detection is sub-optimal, the recent CR-NAS \cite{liang2019computation} reallocates the computation across different resolutions. It is based on the observation that the allocation of computation across different resolutions has a great impact on the Effective Receptive Field (ERF) and affects the detection performance. In BFbox \cite{liu2020bfbox}, a meaningful phenomenon has been observed that the same backbone performs inconsistently between the task of general object detection on COCO \cite{lin2014microsoft} and face detection on WIDER FACE \cite{yang2016wider}, due to the huge gap of scale distribution. Based on this observation, a face-appropriate search space is designed in BFbox \cite{liu2020bfbox}, including a joint optimisation of backbone and neck. In ASFD \cite{zhang2020asfd}, another interesting phenomenon has been observed that some previous Feature Enhance Modules (FAE) perform well in generic object detection but fail in face detection. To this end, a differential architecture search is employed in ASFD \cite{zhang2020asfd} to discover optimised feature enhance modules for efficient multi-scale feature fusion and context enhancement.

Even though the works \cite{liu2020bfbox,zhang2020asfd} have realised the limitation of directly applying general backbone, neck and head settings to face detection, CR-NAS \cite{liang2019computation} only focuses the optimisation on backbone, BFbox \cite{liu2020bfbox} neglects the optimisation of head, and ASFD \cite{zhang2020asfd} only explores the best design for neck.

In this paper, we explore efficient face detection under a fixed VGA resolution (\ie $640\times480$) instead of using large-scale for testing \cite{zhu2020tinaface} to reduce the computation cost. Under this scale setting, most of the faces ($78.93\%$) in WIDER FACE are smaller than $32\times32$ pixels, and thus they are predicted by shallow stages. To obtain more training samples for these shallow stages, we first propose a Sample Redistribution (SR) method through a large cropping strategy. 

Moreover, as the structure of a face detector determines the distribution of computation and is the key in determining its accuracy and efficiency, we further discover principles of computation distribution under different flop regimes. Inspired by \cite{radosavovic2020designing}, we control the degrees of freedom and reduce the search space. More specifically, we randomly sample model architectures with different configurations on backbone (stem and four stages), neck and head. Based on the statistics of these models, we follow \cite{radosavovic2020designing} to compute the empirical bootstrap~\cite{Efron1994} and estimate the likely range in which the best models fall. To further decrease the complexity of the search space, we divide the computation ratio estimation for backbone and the whole detector into two steps.

To sum up, this paper makes the following contributions:
\begin{tight_itemize}
\setlength\itemsep{0em}
\item We explore the efficiency of face detection under VGA resolution, and propose a sample redistribution strategy (SR) that helps to obtain more training samples for shallow stages.

\item We design a simplified search space for computation redistribution across different components (backbone, neck and head) of a face detector. The proposed two-step computation redistribution method can easily gain insight on computation allocation.

\item Extensive experiments conducted on WIDER FACE demonstrate the significantly improved accuracy and efficiency trade-off of the proposed \scrfd across a wide range of compute regimes as shown in Fig. \ref{fig:evelope}. 
\end{tight_itemize}

\section{Related Work}
\noindent{\bf Face Detection.} To deal with extreme variations (\eg scale, pose, illumination and occlusion) in face detection~\cite{yang2016wider}, most of the recent single-shot face detectors focus on improving the anchor sampling/matching or feature enhancement.
SSH~\cite{najibi2017ssh} builds detection modules on different feature maps with a rich receptive field. S$^3$FD~\cite{zhang2017s3fd} introduces an anchor compensation strategy by offsetting anchors for outer faces. PyramidBox~\cite{tang2018pyramidbox} formulates a data-anchor-sampling strategy to increase the proportion of small faces in the training data. DSFD~\cite{li2019dsfd} introduces small faces supervision signals on the backbone, which implicitly boosts the performance of pyramid features. Group sampling~\cite{ming2019group} emphasizes the importance of the ratio for matched and unmatched anchors. RetinaFace~\cite{deng2019retinaface} employs deformable context modules and additional landmark annotations to improve the performance of face detection. HAMBox~\cite{liu2019hambox} finds that many unmatched anchors in the training phase also have strong localization ability and proposes an online high-quality anchor mining strategy to assign high-quality anchors for outer faces. BFbox \cite{liu2020bfbox} employs a single-path one-shot search method \cite{guo2019single} to jointly optimise the backbone and neck for face detector. ASFD \cite{zhang2020asfd} explores a differential architecture search to discover optimised feature enhance modules for efficient multi-scale feature fusion and context enhancement. All these methods are either designed by expert experience or partially optimised on backbone, neck and head. By contrast, we search for computation redistribution across different components (backbone, neck and head) of a face detector across a wide range of compute regimes, which is a global optimisation.

\noindent{\bf Neural Architecture Search.} Given a fixed search space of possible networks, Neural Architecture Search (NAS) automatically finds a good model within the search space. DetNAS \cite{chen2019detnas} adopts the evolution algorithm for the backbone search to boost object detection on COCO \cite{lin2014microsoft}. By contrast, CR-NAS \cite{liang2019computation} reallocates the computation across different stages within the backbone to improve object detection. NAS-FPN \cite{ghiasi2019fpn} uses reinforcement learning to search the proper FPN for general object detection. As there is an obvious distribution gap between COCO \cite{lin2014microsoft} and WIDER FACE \cite{yang2016wider}, the experience in the above methods is not directly applicable for face detection but gives us an inspiration that the backbone, neck and head can be optimised to enhance the performance of face detection. Inspired by RegNet \cite{radosavovic2020designing}, we optimise the computation distribution on backbone, neck and head based on the statistics from a group of randomly sampled models. We successfully reduce the search space and find the stable computation distribution under a particular complex regime, which significantly improves the model's performance.

\section{TinaFace Revisited}

\begin{figure*}[t!]
\centering
\subfigure[Different Testing Scales]{
\label{fig:tinafacetestroc}
\includegraphics[height=0.22\linewidth]{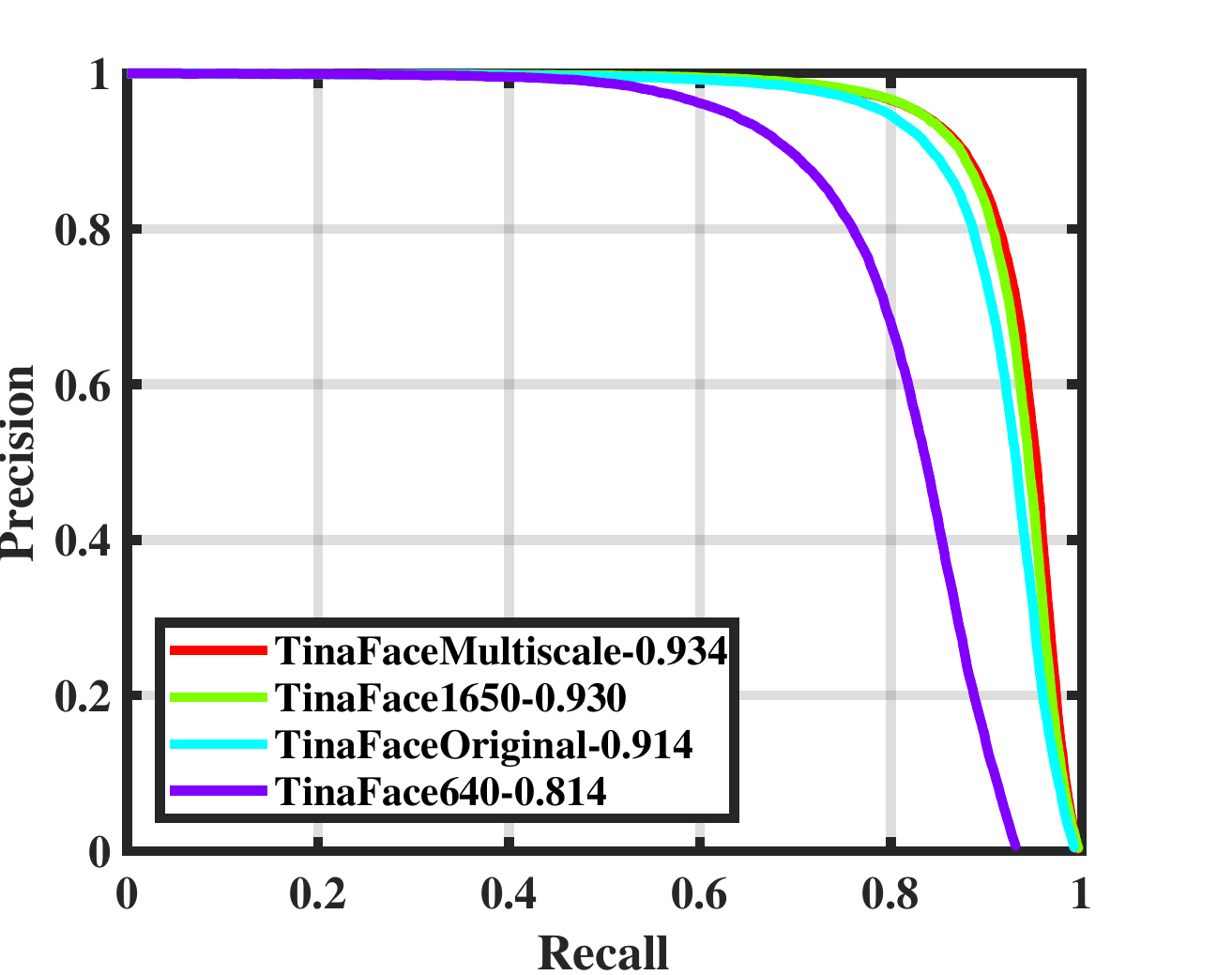}}
\subfigure[Computation Distribution of TinaFace@$640\times480$]{
\label{fig:tinafacecomputationdistribution}
\includegraphics[height=0.22\linewidth]{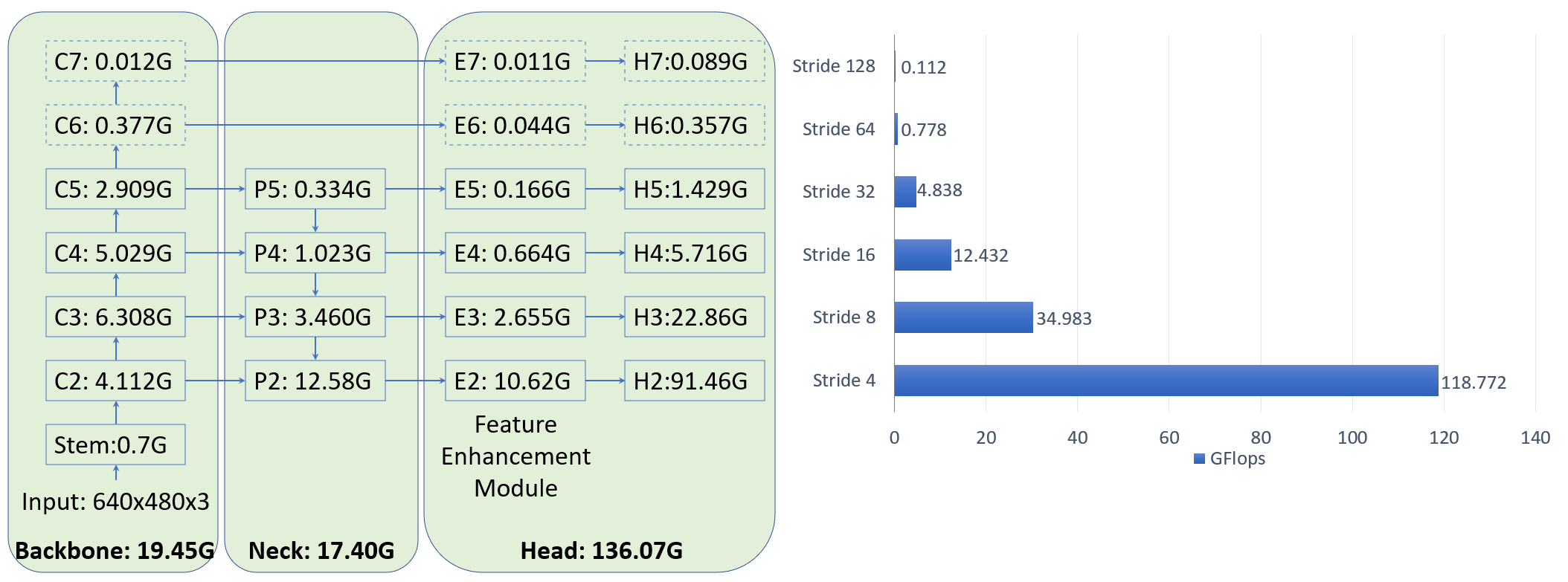}}
\caption{(a) Precision-recall curves of TinaFace-ResNet50 on the WIDER FACE hard validation subset, under different testing scales. (b) Computation distribution of TinaFace on backbone, neck and head with $640\times480$ as the testing scale.}
\vspace{-4mm}
\label{fig:tinafacerevisited}
\end{figure*}

\begin{table}[t]
\small
\centering
\resizebox{\linewidth}{!}{
\begin{tabular}{c|c|c|c}
\hline
Testing Scale & AP & AP@Precision $>98\%$ & \#Flops(G) \\
\hline\hline
Multi-scale   & 0.934 &  0.746  & 42333.64\\
1650          & 0.930 &  0.750  & 1021.82\\
Original(1024)& 0.914 &  0.706  & 508.47\\
640           & 0.814 &  0.539  & 172.95\\
\hline
\end{tabular}}
\caption{Performance and computation comparisons of TinaFace under different testing scales. The average scale of original images is around $882\times1024$.}
\label{tab:tinafacedifscale}
\end{table}

Based on RetinaNet \cite{lin2017feature}, TinaFace \cite{zhu2020tinaface} employs ResNet-50 \cite{he2016deep} as backbone, and Feature Pyramid Network (FPN) \cite{lin2017feature} as neck to construct the feature extractor. For the head design, TinaFace first uses a feature enhancement module on each feature pyramid to learn surrounding context through different receptive fields in the inception block \cite{szegedy2015going}. 
Then, four consecutive $3\times 3$ convolutional layers are appended on each feature pyramid. Focal loss \cite{lin2017focal} is used for the classification branch, DIoU loss \cite{zheng2020distance} for the box regression branch and cross-entropy loss for the IoU prediction branch.

To detect tiny faces, TinaFace tiles anchors of three different scales, over each level of the FPN (\ie $\{2^{4/3}, 2^{5/3},2^{6/3}\} \times \{4, 8, 16, 32, 64, 128\}$, from level $P_2$ to $P_7$). The aspect ratio is set as $1.3$. During training, square patches are cropped from the original image and resized to $640 \times 640$, using a scaling factor randomly sampled from $[0.3, 0.45, 0.6, 0.8, 1.0]$, multiplied by the length of the original image's short edge. During testing, TinaFace employs single scale testing,
when the short and long edge of the image do not surpass $[1100, 1650]$. Otherwise, it employs with short edge scaling at $[500, 800, 1100, 1400, 1700]$, shift with directions $[(0, 0), (0, 1), (1, 0), (1, 1)]$ and horizontal filp.

As shown in Fig.~\ref{fig:tinafacetestroc} and Tab.~\ref{tab:tinafacedifscale}, we compare the performance of TinaFace under different testing scales. For the multi-scale testing, TinaFace achieves an impressive AP of $93.4\%$, which is the current best performance on the WIDER FACE leader-board. For large single-scale testing ($1650$), the AP slightly drops at $93.0\%$ but the computation significantly decreases to $1021.82$ Gflops. On the original scale (\textasciitilde$1024$), the performance of TinaFace is still very high, obtaining an AP of $91.4\%$ with $508.47$ Gflops. Moreover, when the testing scale decreases to VGA level ($640$), the AP significantly reduces to $81.4\%$, with the computation further decreasing at $172.95$ Gflops.

In Fig. \ref{fig:tinafacecomputationdistribution}, we illustrate the computation distribution of TinaFace on the backbone, neck and head components with a testing scale of $640$. From the view of different scales of the feature pyramid, the majority of the computational costs (about $68\%$) are from stride 4, as the resolution of the feature map is quite large ($120\times160$). From the view of the different components of the face detector, most of the computational costs (about $79\%$) are from the head, since the backbone structure is directly borrowed from the ImageNet classification task \cite{deng2009imagenet}, without any modification.

Even though TinaFace achieves state-of-the-art performance on tiny face detection, the heavy computational cost renders it unsuitable for real-time applications. In addition, real-world face detection systems always require high precision (\eg $>98\%$), in order to avoid frequent false alarms. As shown in Tab. \ref{tab:tinafacedifscale}, the APs of TinaFace significantly drop, when the threshold of detection scores is increased to meet the requirement of high precision. 

\section{Methodology}

Based on the above analysis of TinaFace, and the following meticulous experimentation, we propose the following efficiency improvements on the design of face detection, conditioned on (1) the testing scale bounded at the VGA resolution ($640$), and (2) there is no anchor tiled on the feature map of stride 4. In particular, we tile anchors of $\{16, 32\}$ at the feature map of stride 8, anchors of $\{64, 128\}$ at stride 16, and anchors of $\{256, 512\}$ at stride 32. Since our testing scale is smaller, most of the faces will be predicted on stride 8. Therefore, we first investigate the redistribution of positive training samples across different scales of feature maps (Sec.~\ref{sec:sample_reallocation}). Then, we explore the computation redistribution across different scales of feature maps, as well as across different components (\ie backbone, neck and head), given a pre-defined computation budget (Sec.~\ref{sec:computation_redistribution}).

\subsection{Sample Reallocation}
\label{sec:sample_reallocation}
\begin{figure}[t!]
\centering
\includegraphics[width=0.25\textwidth]{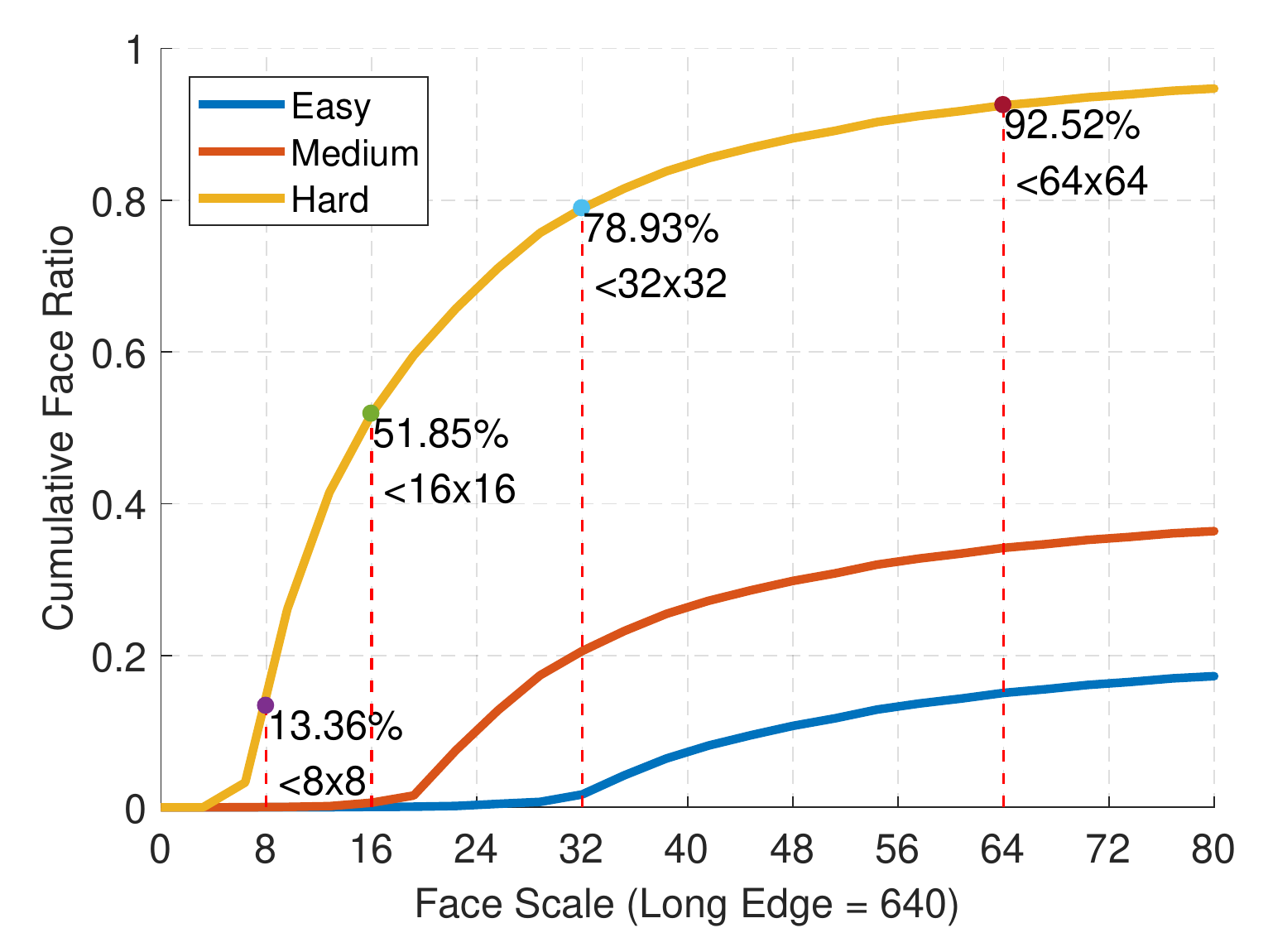}
\caption{Cumulative face scale distribution on the WIDER FACE validation dataset (Easy $\subset$ Medium $\subset$ Hard). When the long edge is fixed as $640$ pixels, most of the easy faces are larger than $32 \times 32$, and most of the medium faces are larger than $16 \times 16$. For the hard track, 78.93\% faces are smaller than $32 \times 32$, 51.85\% faces are smaller than $16 \times 16$, and 13.36\% faces are smaller than $8 \times 8$.}
\vspace{-4mm}
\label{fig:cumulativewiderval}
\end{figure}

\begin{figure}[t!]
\centering
\subfigure[Ground-truth Distribution]{
\label{fig:gtdis}
\includegraphics[width=0.23\textwidth]{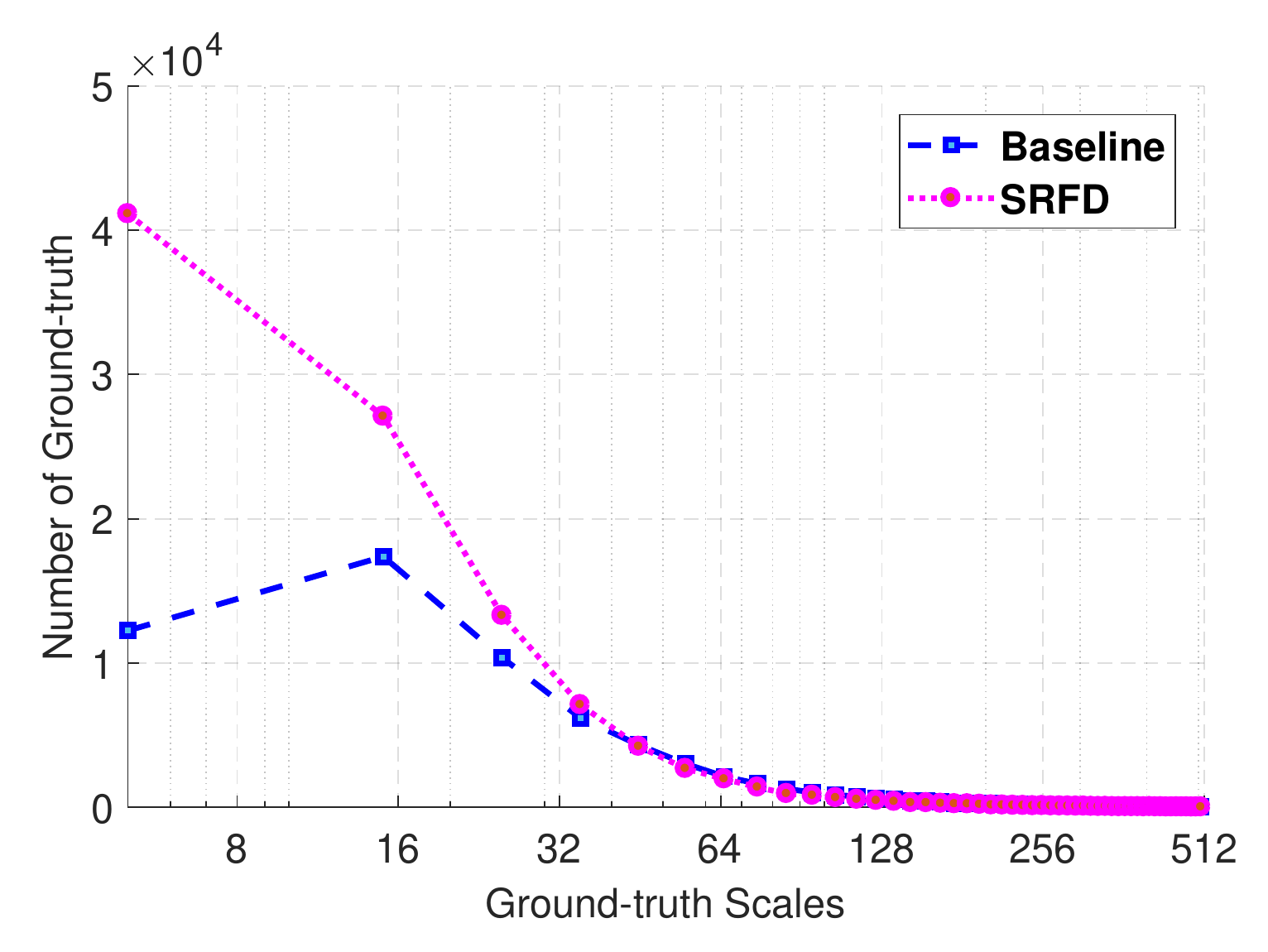}}
\subfigure[Positive Anchor Distribution]{
\label{fig:posandis}
\includegraphics[width=0.23\textwidth]{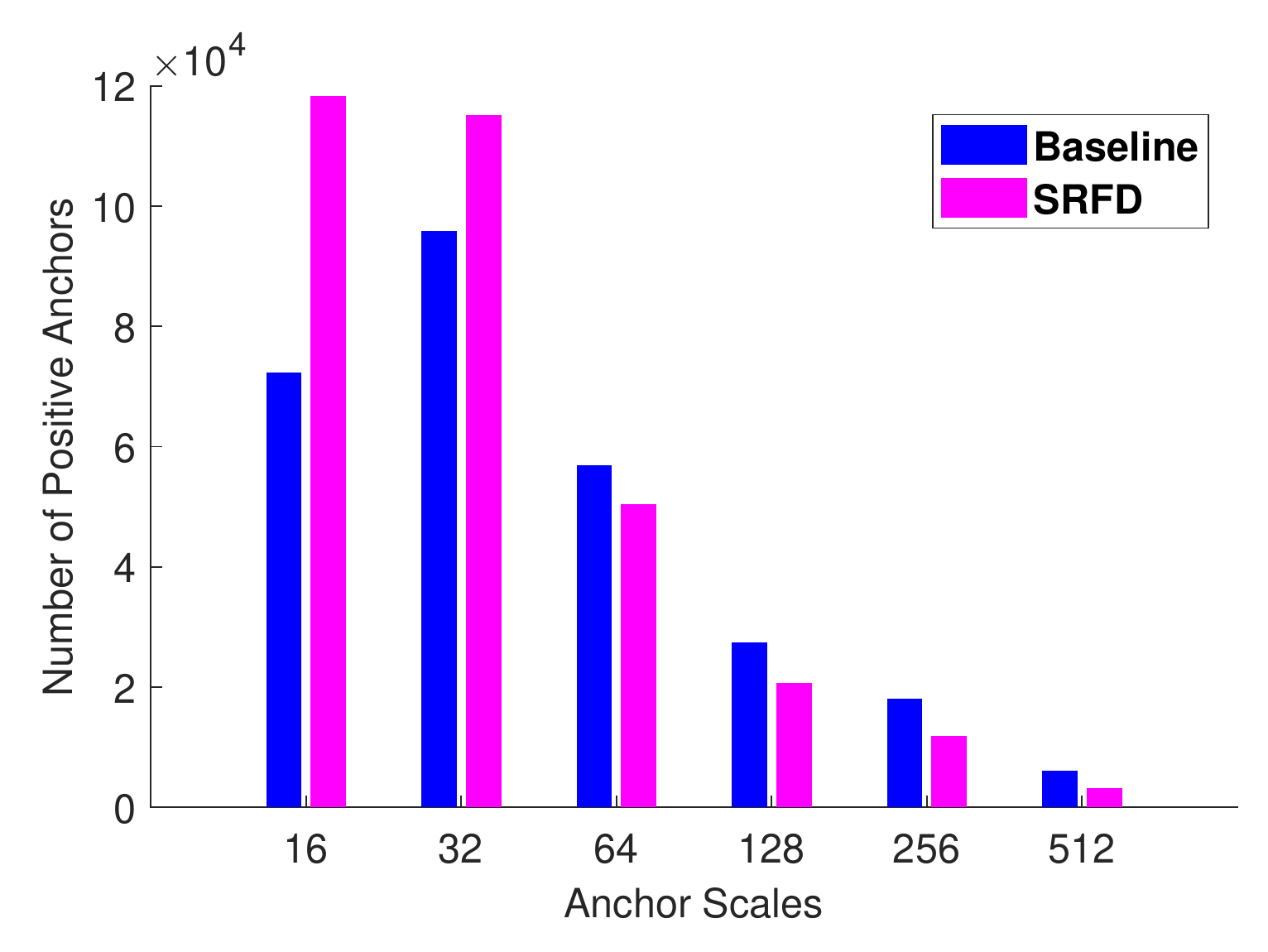}}
\caption{Ground-truth and positive anchor distribution within one epoch. The baseline method employs a random size from the set $[0.3, 1.0]$, while our method uses a random size from the set $[0.3, 2.0]$. The number of small faces ($<32\times32$) significantly increases after the large cropping strategy is used.}
\vspace{-4mm}
\label{fig:samplereallocation}
\end{figure}

We find that the feature map of stride 8 is most important in our setting.
This is evident in Fig.~\ref{fig:cumulativewiderval}, where we draw the cumulative face scale distribution on the WIDER FACE validation dataset.
When the test scale is fixed as $640$ pixels, 78.93\% faces are smaller than $32 \times 32$.

During training data augmentation, square patches are cropped from the original images with a random size from the set $[0.3, 1.0]$ of the short edge of the original images. To generate more positive samples for stride 8, we enlarge the random size range from $[0.3, 1.0]$ to $[0.3, 2.0]$. When the crop box is beyond the original image, average RGB values fill the missing pixels. As illustrated in Fig.~\ref{fig:gtdis}, there are more faces below the scale of $32$ after the proposed large cropping strategy is used. Moreover, even though there will be more extremely tiny faces (\eg $<4\times4$) under the large cropping strategy, these ground-truth faces will be neglected during training due to unsuccessful anchor matching. As shown in Fig.~\ref{fig:posandis}, positive anchors within one epoch significantly increase from 72.3K to 118.3K at the scale of $16$, and increase from 95.9K to 115.1K at the scale of $32$. With more training samples redistributed to the small scale, the branch to detect tiny faces can be trained more adequately. 

\subsection{Computation Redistribution}
\label{sec:computation_redistribution}

Directly utilising the backbone of a classification network for scale-specific face detection can be sub-optimal. Therefore, we employ network structure search \cite{radosavovic2020designing} to reallocate the computation on the backbone, neck and head, under a wide range of flop regimes. We apply our search method on RetinaNet \cite{lin2017feature}, with ResNet \cite{he2016deep} as backbone, Path Aggregation Feature Pyramid Network (PAFPN) \cite{liu2018path} as the neck and stacked $3\times3$ convolutional layers for the head. While the general structure is simple, the total number of possible networks in the search space is unwieldy. In the first step, we explore the reallocation of the computation within the backbone parts (\ie stem, C2, C3, C4, and C5), while fixing the neck and head components. Based on the optimised computation distribution on the backbone that we find, we further explore the reallocation of the computation across the backbone, neck and head. By optimising in both manners, we achieve the final optimised network design for face detection.

\noindent{\bf Computation search space reduction.}
Our goal is to design better networks for efficient face detection, by redistributing the computation. Given a fixed computation cost, as well as the face scale distribution presented in Fig.~\ref{fig:cumulativewiderval}, we explore the relationship between computation distribution and performance from populations of models. 

Following RegNet \cite{radosavovic2020designing}, we explore the structures of face detectors, assuming fixed standard network blocks (\ie, basic residual or bottleneck blocks with a fixed bottleneck ratio of $4$). In our case, the structure of a face detector includes: (1) the \textit{backbone stem}, three $3\times3$ convolutional layers with $w_0$ output channels \cite{he2019bag}, (2) the \textit{backbone body}, four stages operating at progressively reduced resolution, with each stage consisting of a sequence of identical blocks. For each stage $i$, the degrees of freedom include the number of blocks $d_i$ (\ie network depth) and the block width $w_i$ (\ie number of channels). The structure of the face detector also includes: (3) the \textit{neck}, a multi-scale feature aggregation module by a top-down path and a bottom-up path with $n_i$ channels \cite{liu2018path}, (4) the \textit{head}, with $h_i$ channels of $m$ blocks to predict face scores and regress face boxes. 

As the channel number of the stem is equal to the block width of the first residual block in C2, the degree of freedom of the stem can be merged into $w_1$. In addition, we employ a shared head design for three-scale of feature maps and fix the channel number for all $3\times3$ convolutional layers within the heads. Therefore, we reduce the degrees of freedom to three within our neck and head design: (1) output channel number $n$ for neck, (2) output channel number $h$ for head, and (3) the number of $3\times3$ convolutional layers $m$. We perform uniform sampling of $n \leq 256$, $h\leq 256$, and $m\leq 6$ (both $n$ and $h$ are divisible by 8).

The backbone search space has 8 degrees of freedom as there are 4 stages and each stage $i$ has 2 parameters: the number of blocks $d_i$ and block width $w_i$. Following RegNet \cite{radosavovic2020designing}, we perform uniform sampling of $d_i\leq 24$ and $w_i\leq 512$ ($w_i$ is divisible by 8). 
As state-of-the-art backbones have increasing widths \cite{radosavovic2020designing}, we also shrink the search space constrained to the principle of $w_{i+1}\ge w_i$.

With the above simplifications, our search space becomes more straightforward. We repeat the random sampling in our search space until we obtain $320$ models in our target complexity regime, and train each model on the WIDER FACE training set for $80$ epochs. Then, we test the AP of each model on the validation set. 
Based on these $320$ pairs of model statistics $(x_i, AP_i)$, where $x_i$ is the computation ratio of a particular component and $AP_i$ the corresponding performance, we follow \cite{radosavovic2020designing} to compute the empirical bootstrap~\cite{Efron1994} to estimate the likely range in which the best models fall. 

Finally, to further decrease the complexity of search space, we divide our network structure search into the following two steps:
\begin{tight_itemize}
\item \scrfdv{1}: search the computational distribution for the backbone only, while fixing the settings of neck and head to the default configuration.
\item \scrfdv{2}: search the computational distribution over the whole face detector (\ie backbone, neck and head), with the computational distribution within the backbone following the optimised \scrfdv{1}. 
\end{tight_itemize}
Here, we use \scrfd constrained to 2.5 Gflops (\scrfdf{2.5}) as an example to illustrate our two-step searching strategy.

\begin{figure*}
\centering
\subfigure[Stem $\sim (10\%, 20\%)$]{
\label{fig:stem}
\includegraphics[width=0.19\linewidth]{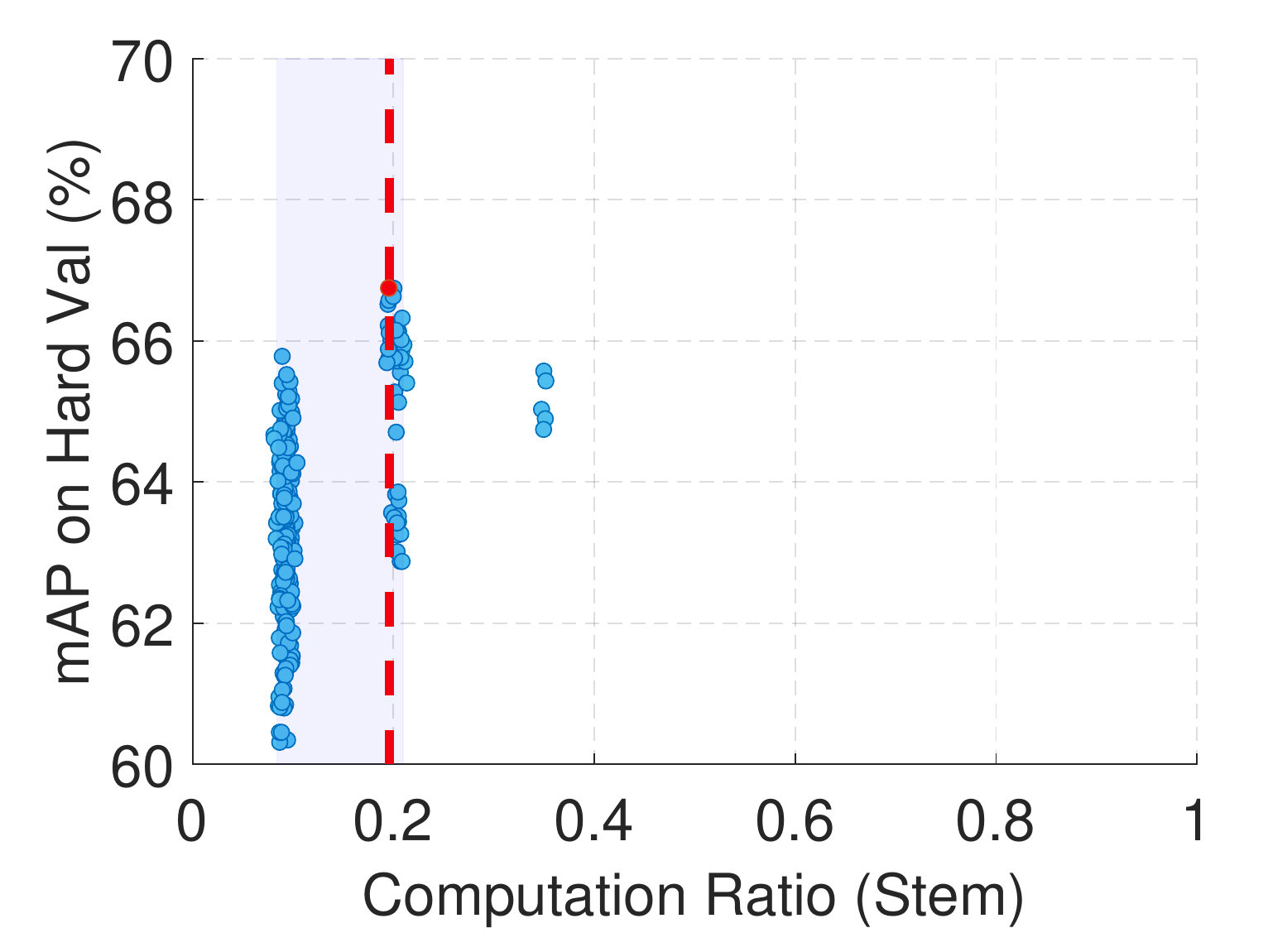}}
\subfigure[C2 $\sim (24\%, 39\%)$]{
\label{fig:c2}
\includegraphics[width=0.19\linewidth]{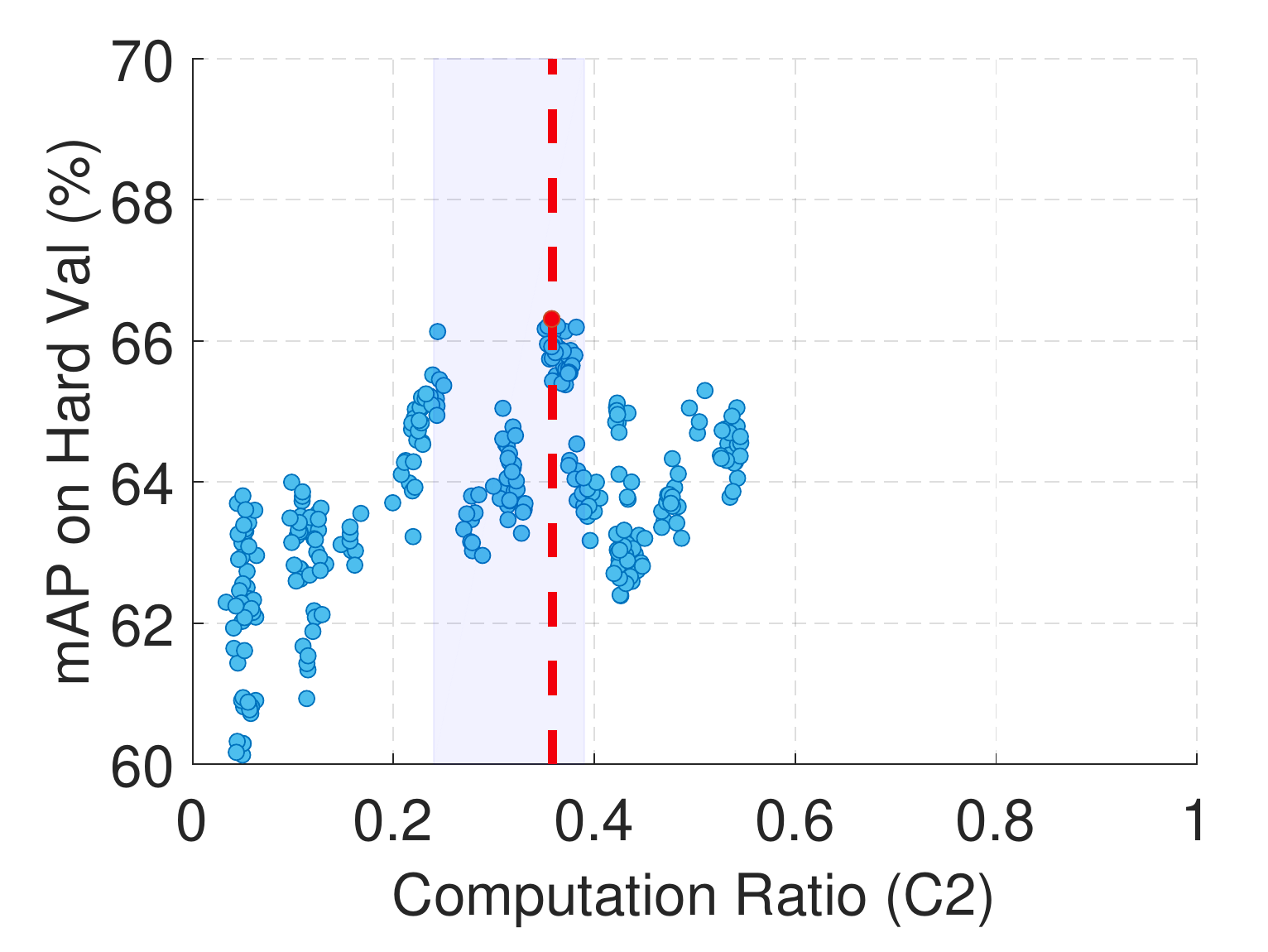}}
\subfigure[C3 $\sim (26\%, 47\%)$]{
\label{fig:c3}
\includegraphics[width=0.19\linewidth]{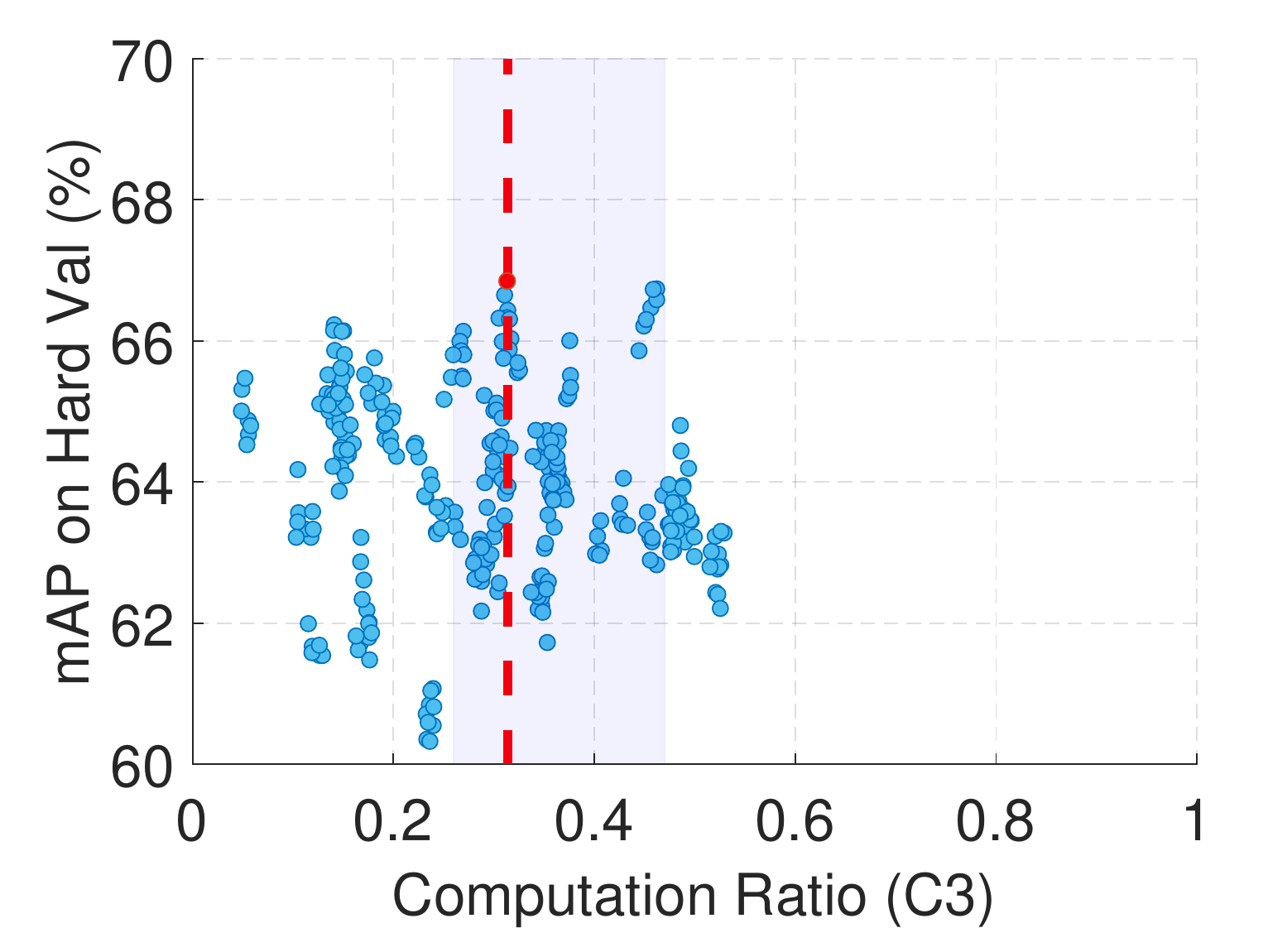}}
\subfigure[C4 $\sim (4\%, 15\%)$]{
\label{fig:c4}
\includegraphics[width=0.19\linewidth]{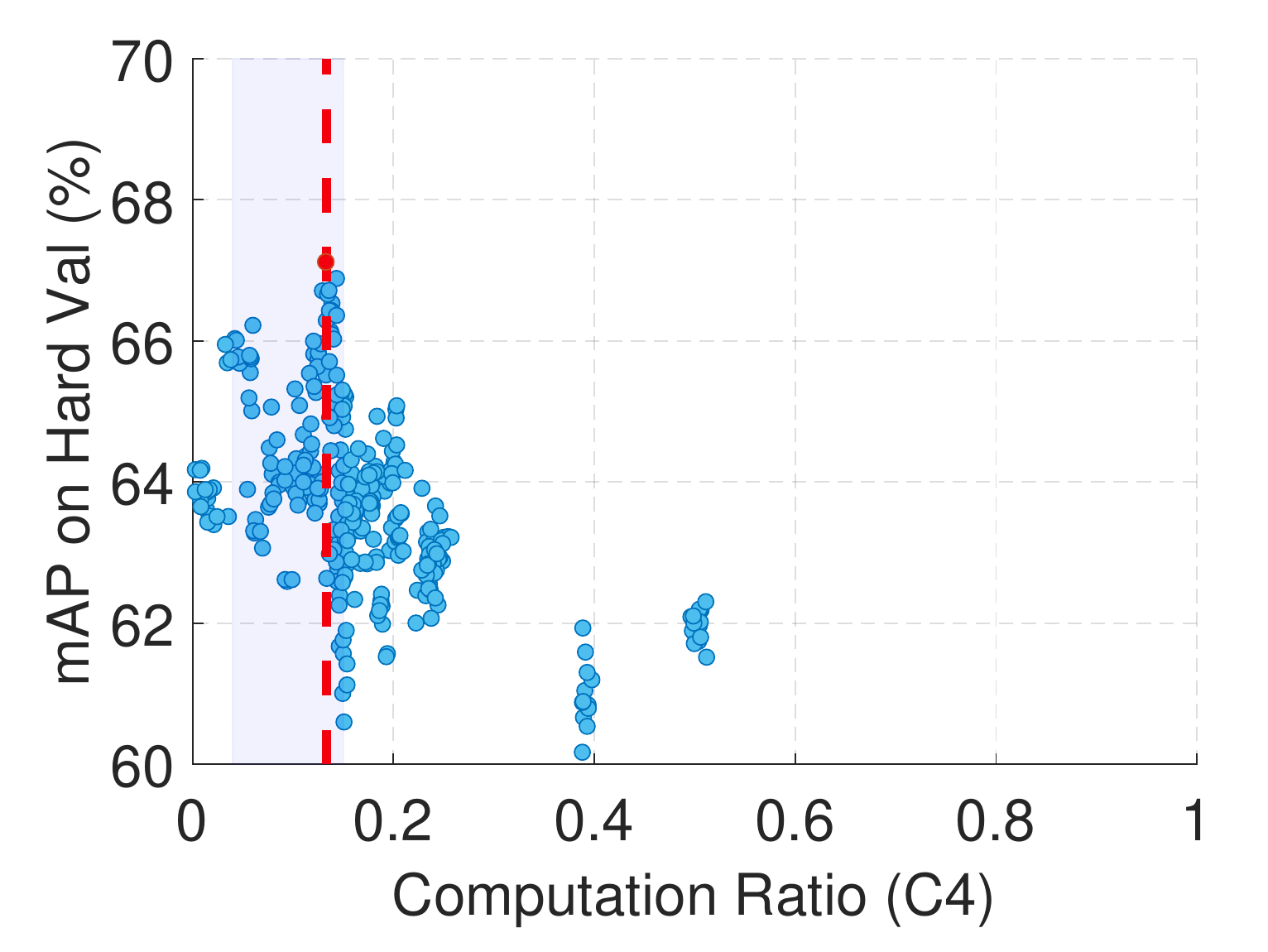}}
\subfigure[C5 $\sim (1\%, 16\%)$]{
\label{fig:c5}
\includegraphics[width=0.19\linewidth]{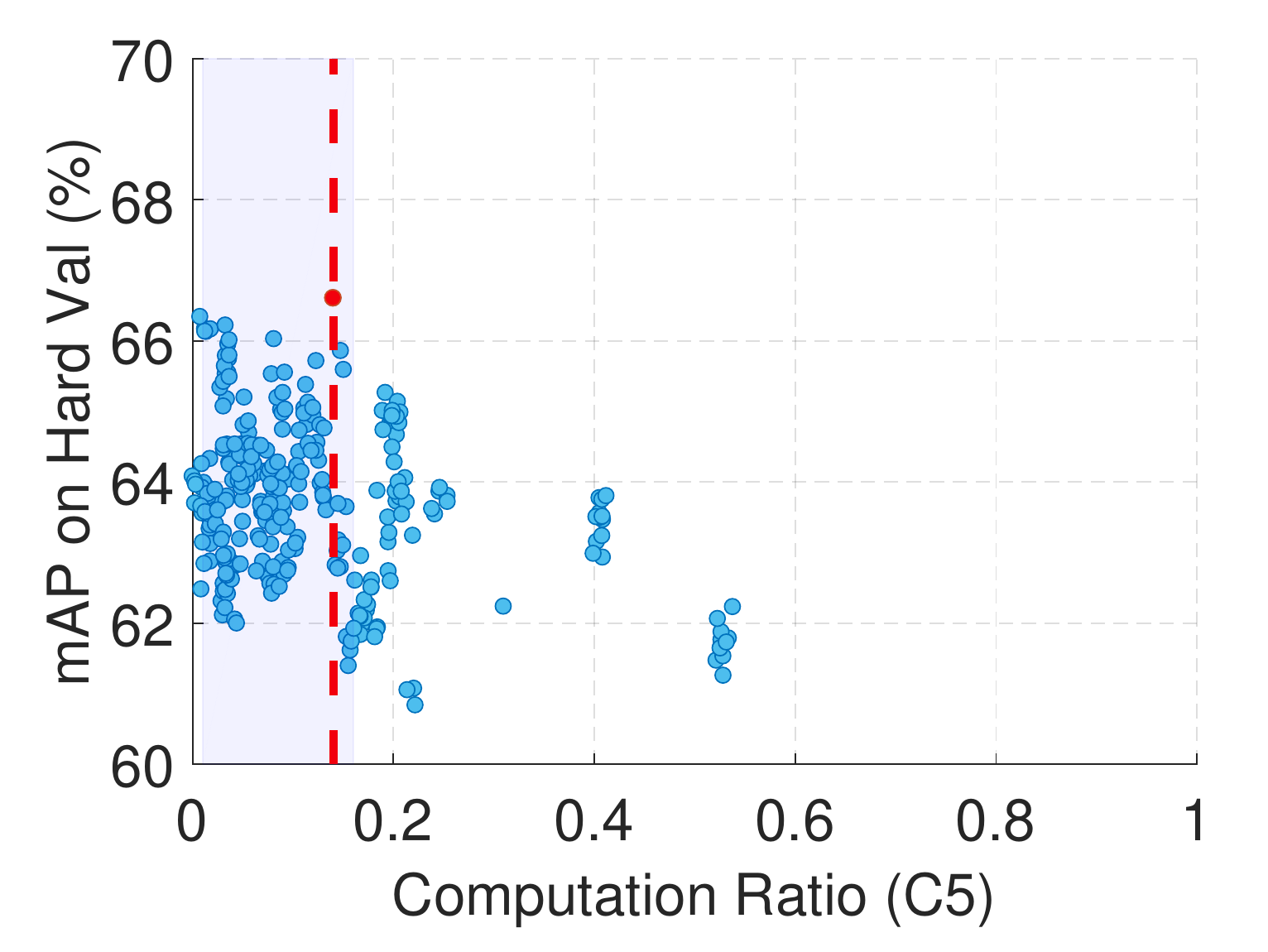}}
\caption{Computation redistribution on the backbone (stem, C2, C3, C4 and C5) with fixed neck and head under the constraint of 2.5 Gflops. 
For each component within the backbone, the range of computation ratio in which the best models may fall is estimated by the empirical bootstrap. }
\label{fig:computationoptimizationbackbone}
\end{figure*}

\begin{figure}[h]
\centering
\subfigure[Stem+C2+C3 $\sim (72\%, 91\%)$]{
\label{fig:Stem+C2+C3}
\includegraphics[width=0.45\linewidth]{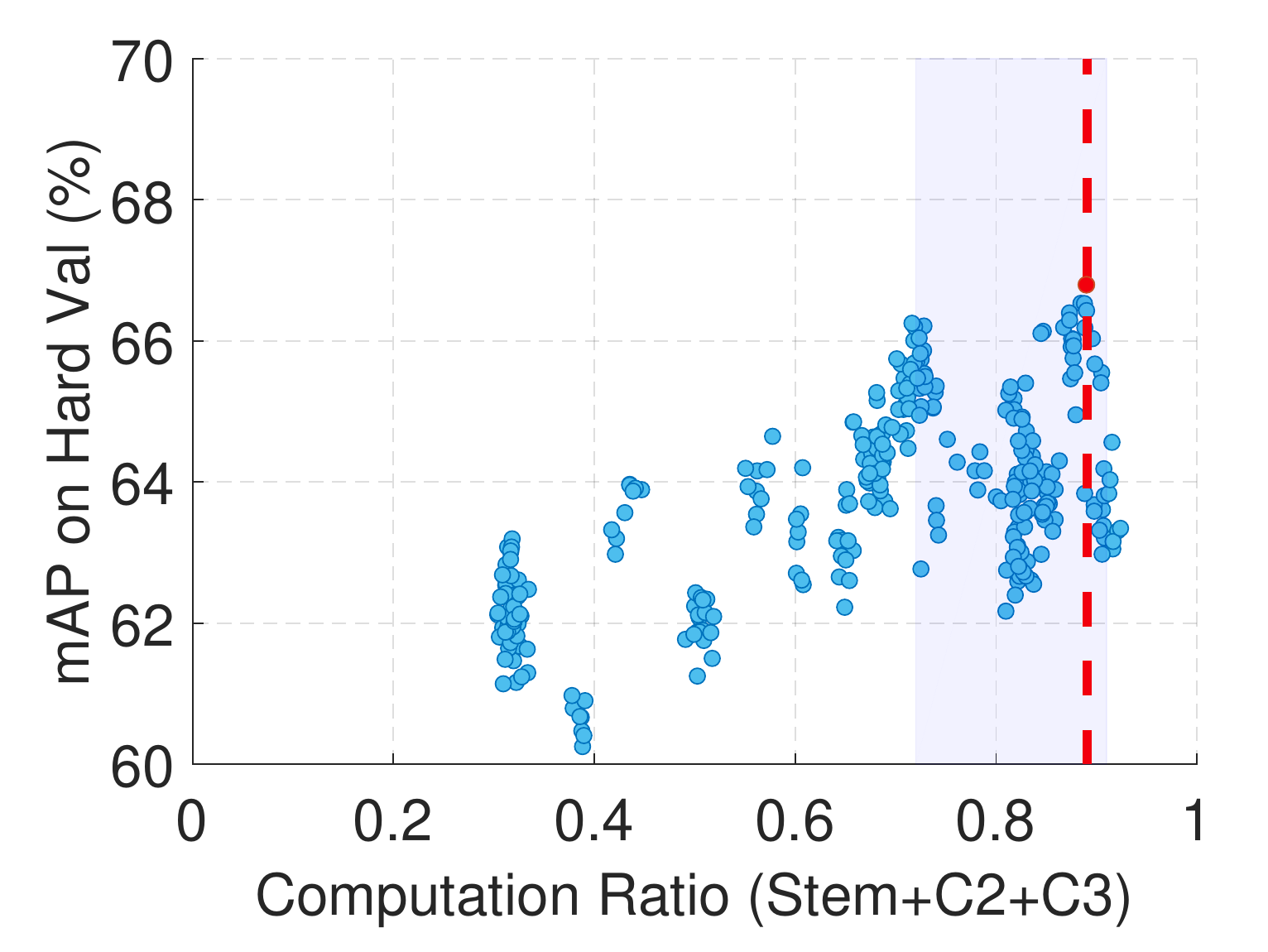}}
\subfigure[C4+C5 $\sim (9\%, 28\%)$]{
\label{fig:C4+C5}
\includegraphics[width=0.45\linewidth]{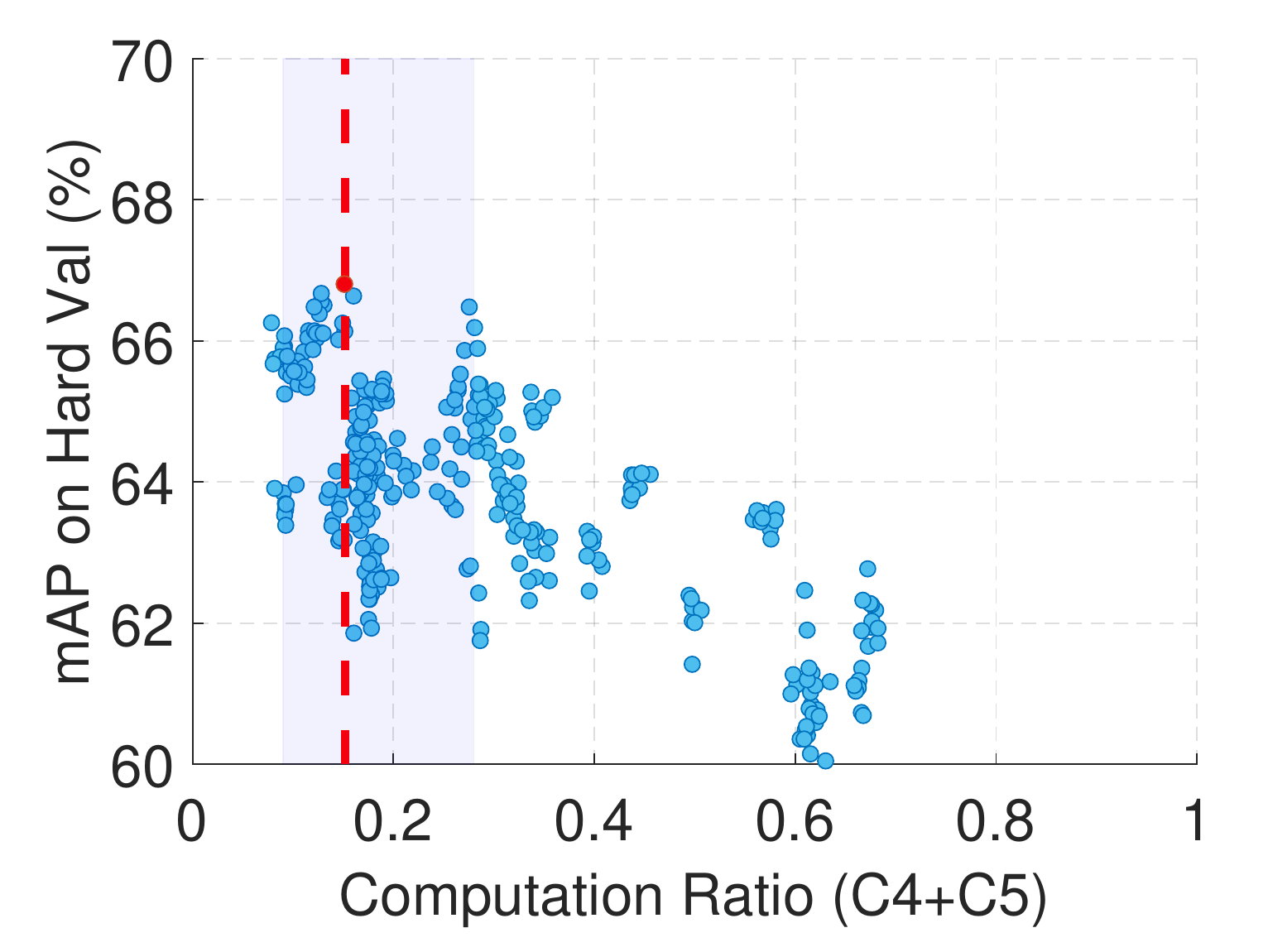}}
\caption{Computation redistribution between the shallow and deep stages of the backbone under the constraint of 2.5 Gflops.}
\label{fig:computationoptimizationlowhigh}
\end{figure}

\noindent{\bf Computation redistribution on backbone.}
Since the backbone performs the bulk of the computation, we first focus on the structure of the backbone, which is central in determining the network's computational cost and accuracy. For \scrfdvf{1}{2.5}, we fix the output channel of the neck at $32$ and use two stacked $3\times3$ convolutions with $96$ output channels. As the neck and head configurations do not change in the whole search process of \scrfdv{1}, we can easily find the best computation distribution of the backbone. As described in Fig.~\ref{fig:computationoptimizationbackbone}, we show the distribution of $320$ model APs (on the WIDER FACE hard validation set) versus the computation ratio over each component (\ie stem, C2, C3, C4 and C5) of backbone. After applying an empirical bootstrap~\cite{Efron1994}, a clear trend emerges showing that the backbone computation is reallocated to the shallow stages (\ie C2 and C3). In Fig.~\ref{fig:computationoptimizationlowhigh}, we present the computation ratio between the shallow (\ie stem, C2, and C3) and deep stages (\ie C4 and C5) of the backbone. Based on the observation from these search results, it is obvious that around $80\%$ of the computation is reallocated to the shallow stages.

\begin{figure}[t!]
\centering
\subfigure[2.5GF]{
\label{fig:25gflops}
\includegraphics[width=0.225\textwidth]{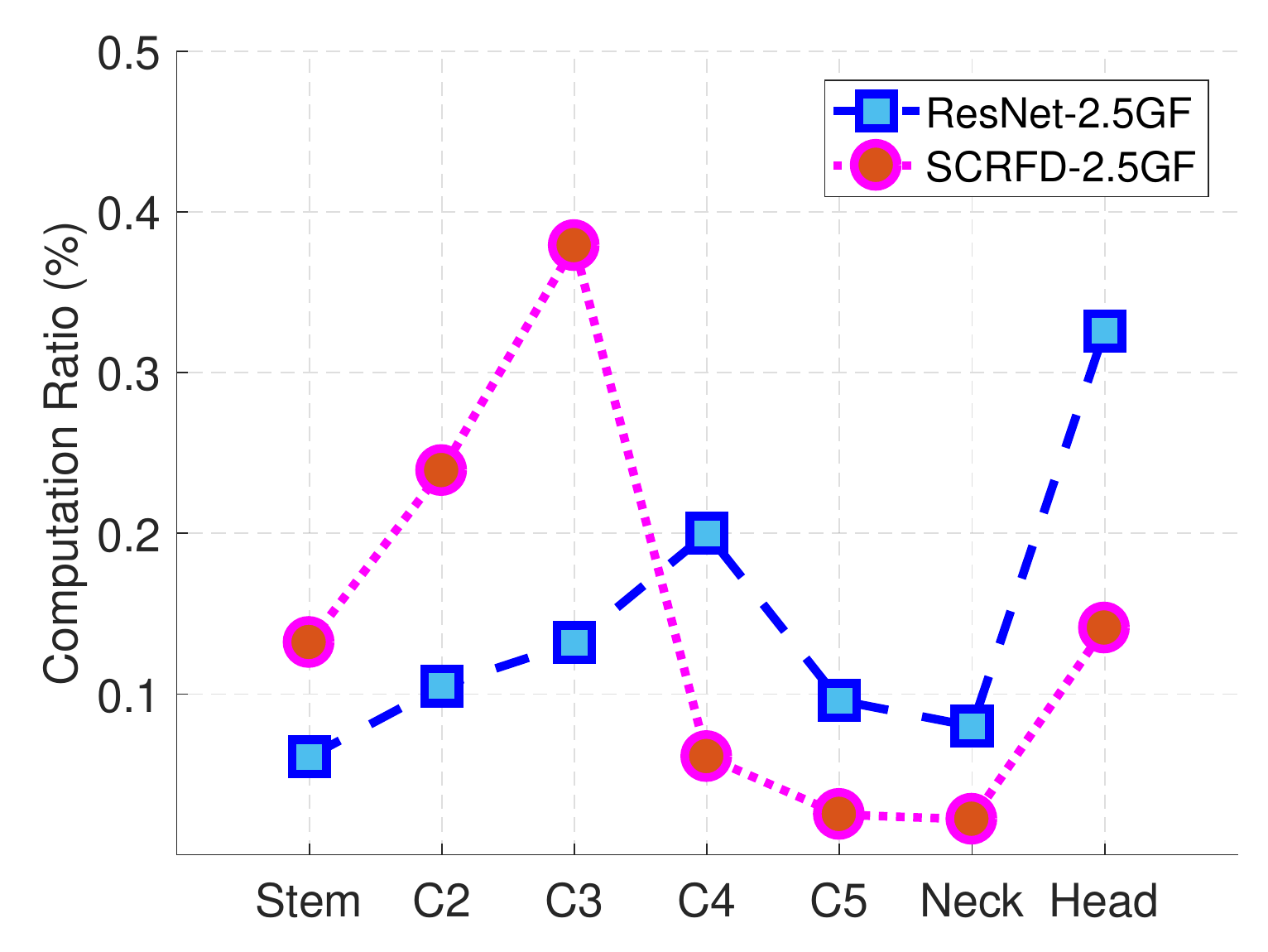}}
\subfigure[\scrfdf{2.5}]{
\label{fig:25gflopscf}
\includegraphics[width=0.225\textwidth]{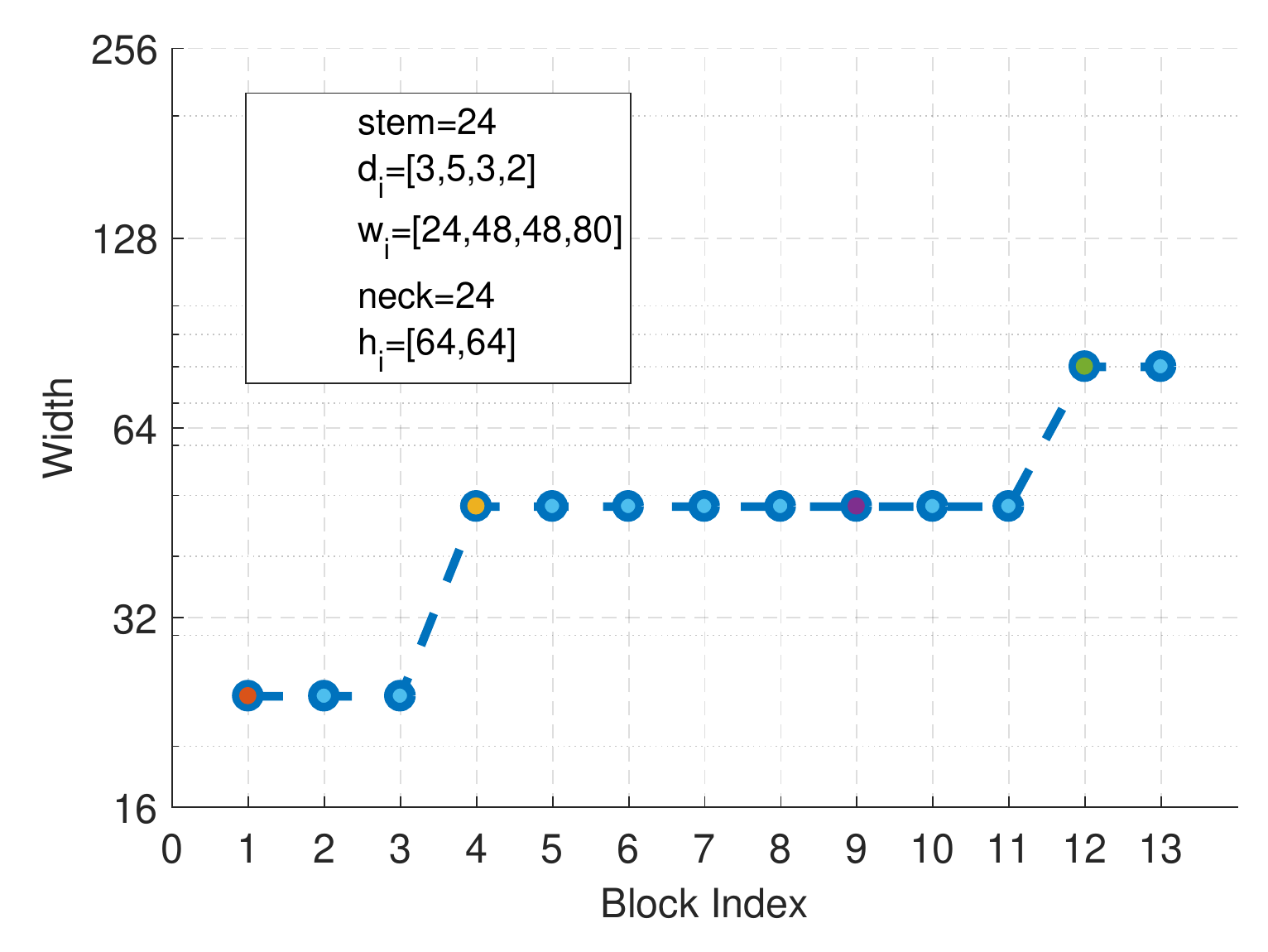}}
\subfigure[10GF]{
\label{fig:10gflops}
\includegraphics[width=0.225\textwidth]{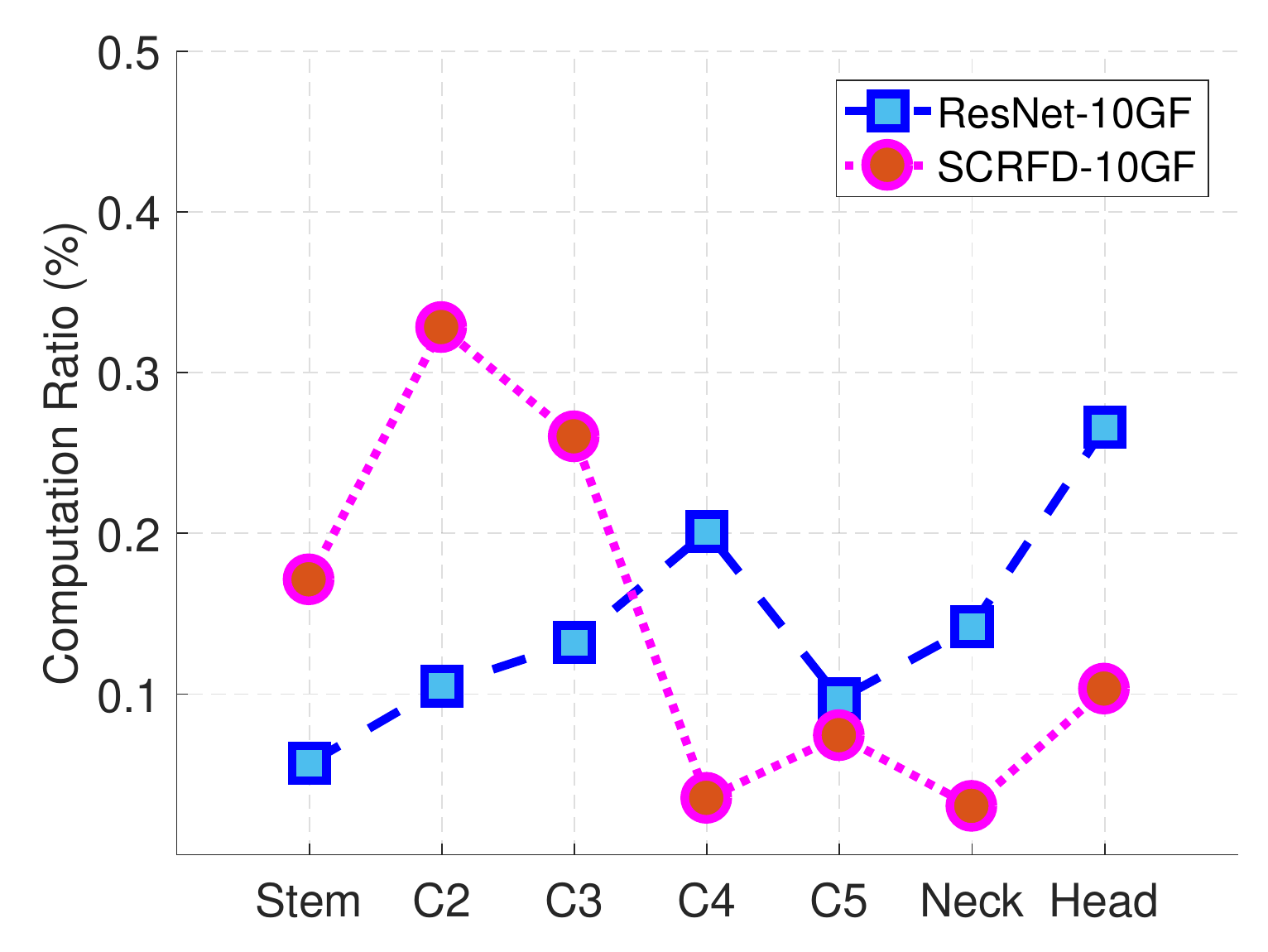}}
\subfigure[\scrfdf{10}]{
\label{fig:10gflopscf}
\includegraphics[width=0.225\textwidth]{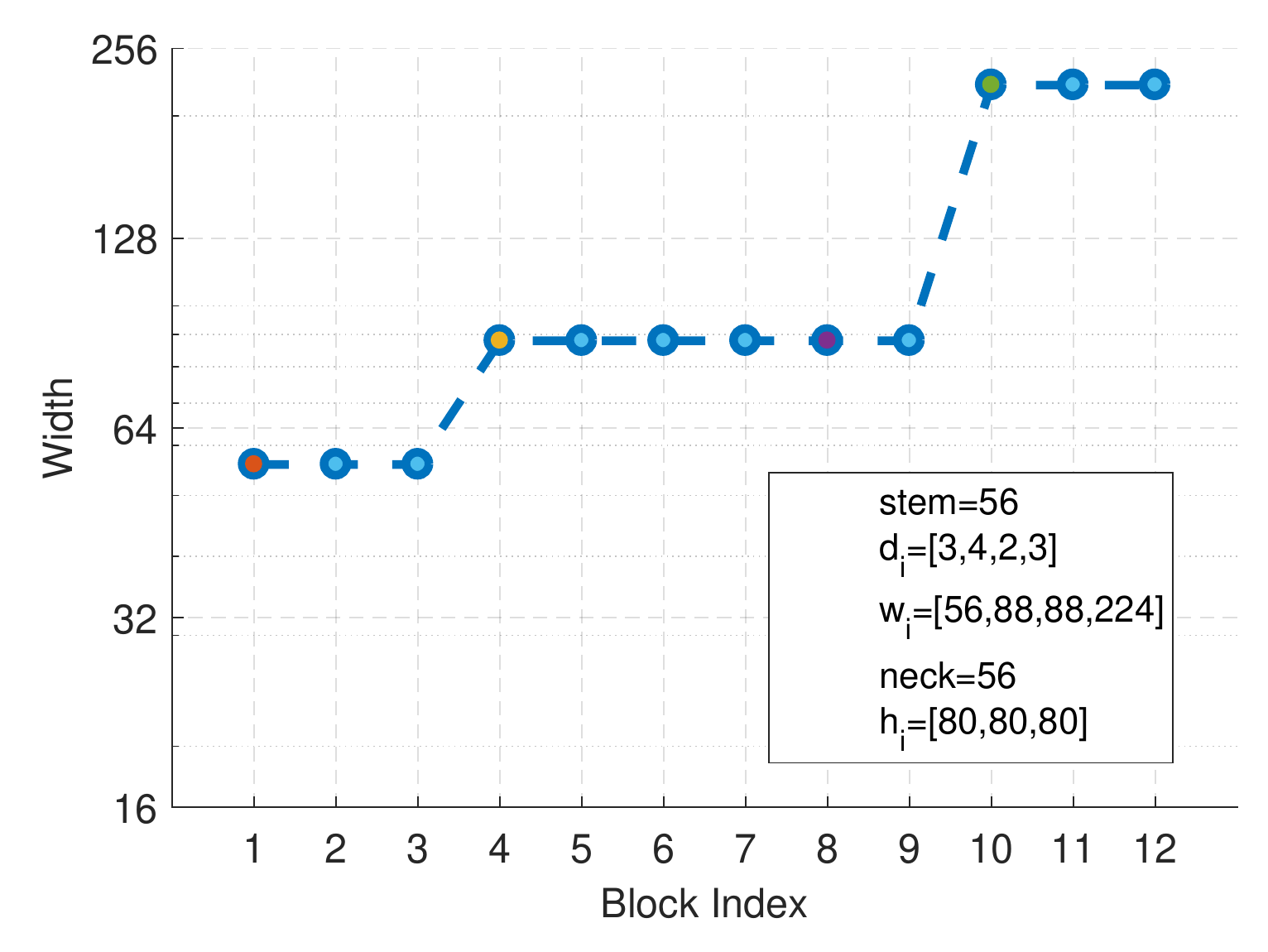}}
\subfigure[34GF]{
\label{fig:34gflops}
\includegraphics[width=0.225\textwidth]{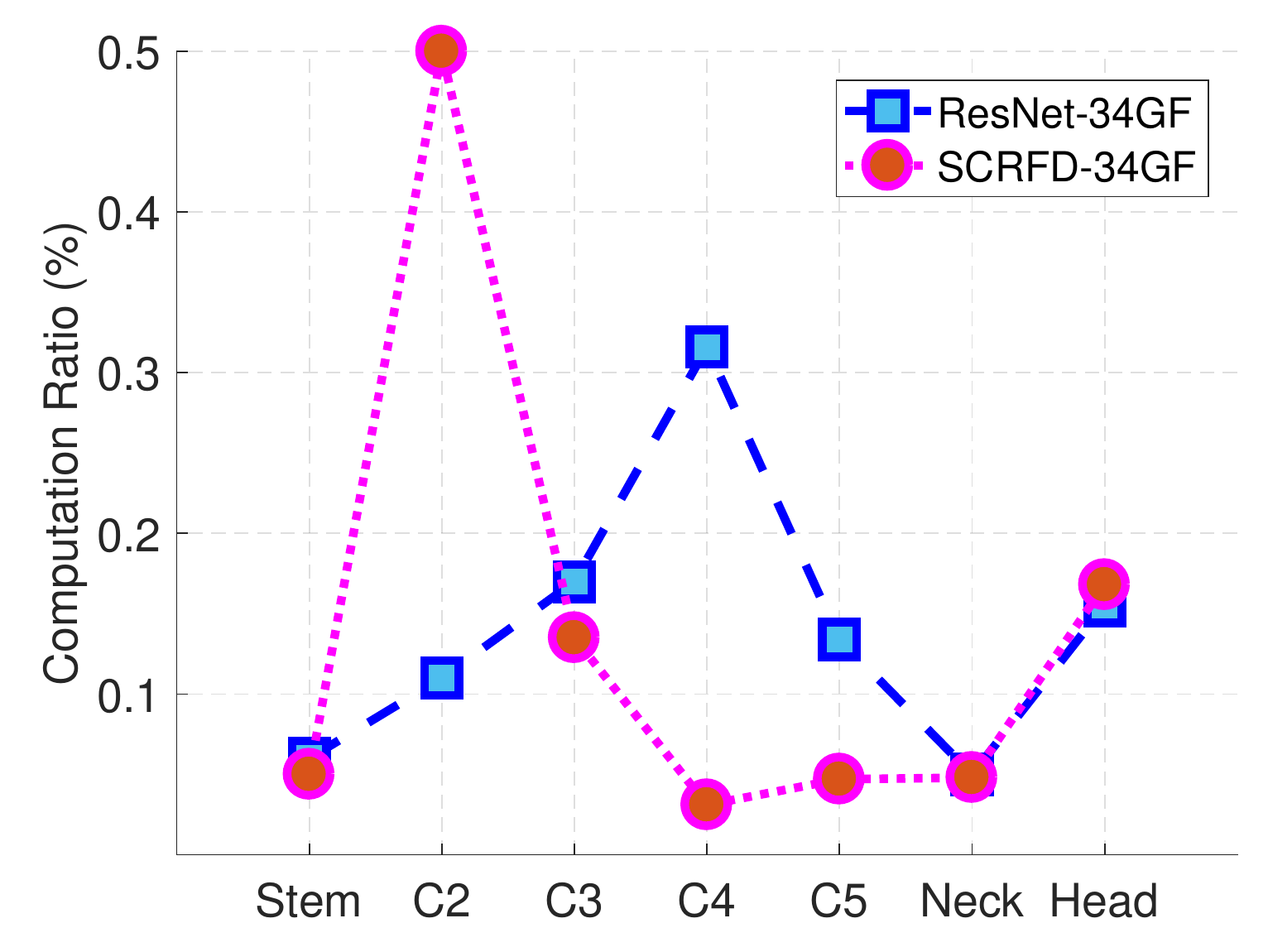}}
\subfigure[\scrfdf{34}]{
\label{fig:34gflopscf}
\includegraphics[width=0.225\textwidth]{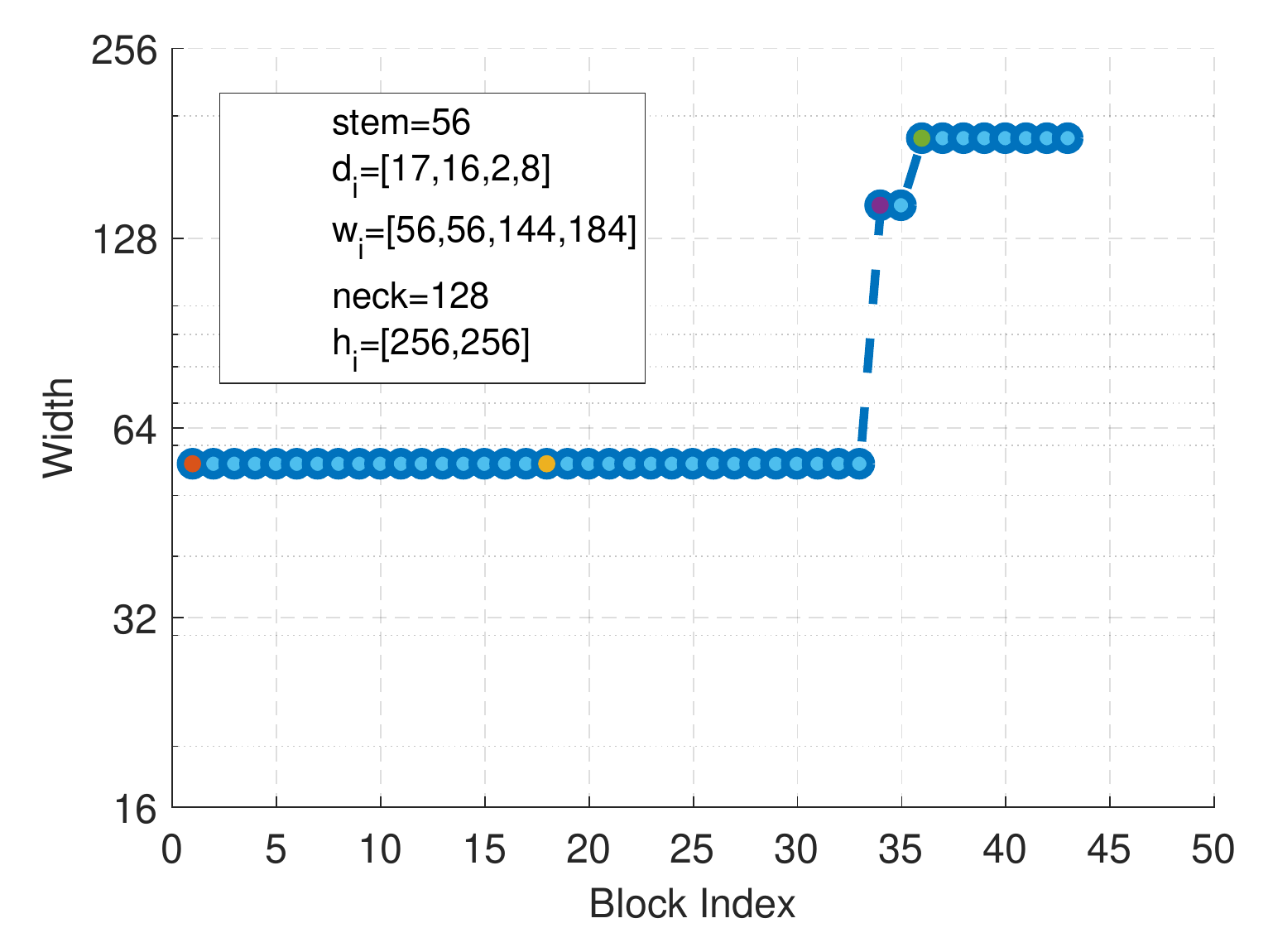}}
\caption{Computation redistribution (left column) and the searched network structures (right column) under different computation constraints (2.5GF, 10GF, and 34GF). Network diagram legends in the second row contain all information required to implement the \scrfd models that we have optimised the computation across stages and components.}
\vspace{-0.08cm}
\label{fig:resnetscalable}
\end{figure}

\begin{figure}[t!]
\centering
\subfigure[0.5GF]{
\label{fig:05gflops}
\includegraphics[width=0.225\textwidth]{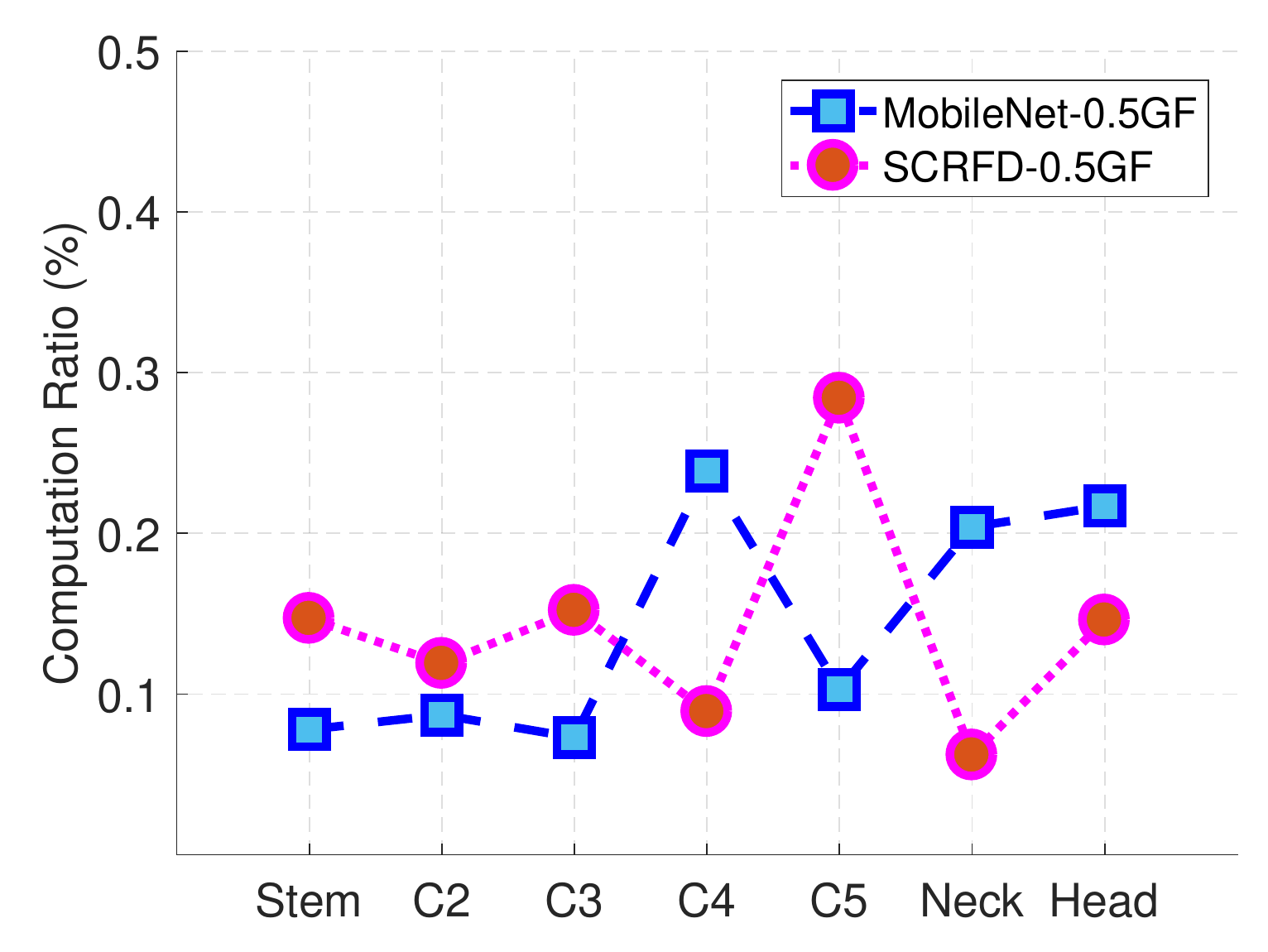}}
\subfigure[\scrfdf{0.5}]{
\label{fig:05gflopscf}
\includegraphics[width=0.225\textwidth]{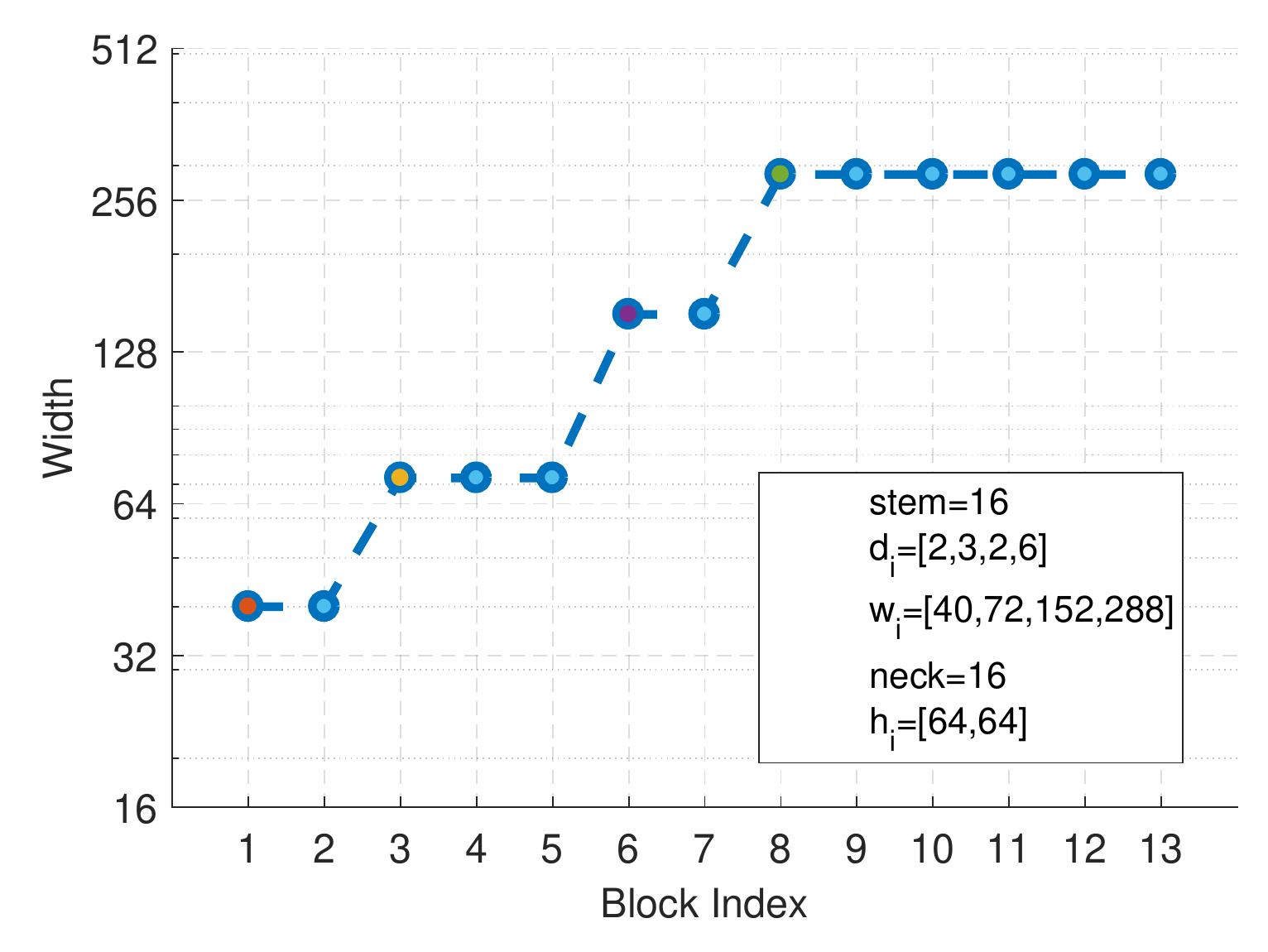}}
\caption{Computation redistribution (left) and the searched network structures (right) for the mobile regime (0.5 Gflops).}
\vspace{-4mm}
\label{fig:mobilenetscalable}
\end{figure}

\begin{figure*}[h]
\footnotesize
\centering
\subfigure[\scriptsize{Backbone $\sim (67\%, 88\%)$}]{
\label{fig:Backbone}
\includegraphics[width=0.18\textwidth]{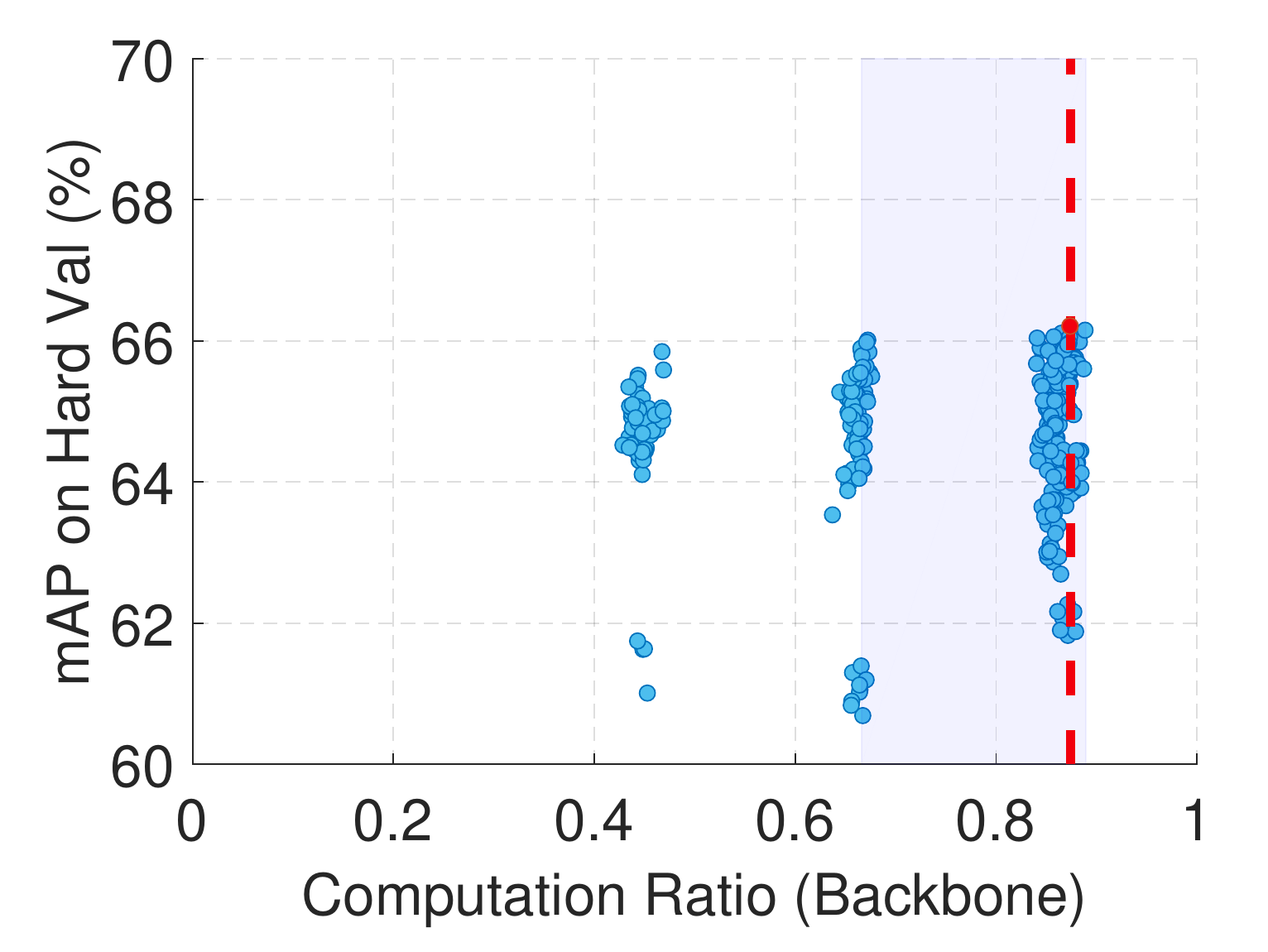}}
\subfigure[Neck $\sim (1\%, 7\%)$]{
\label{fig:Neck}
\includegraphics[width=0.18\textwidth]{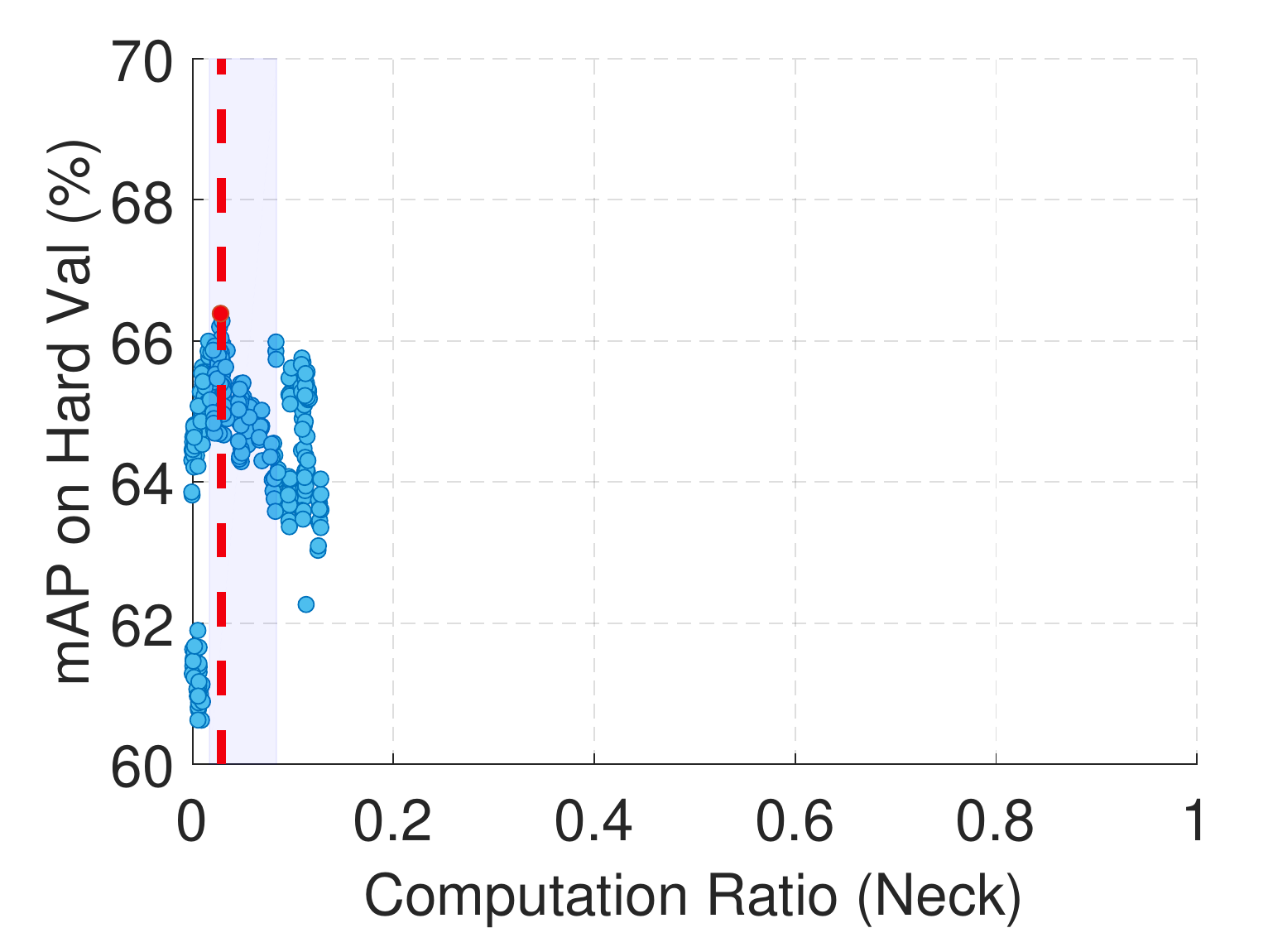}}
\subfigure[Head $\sim (10\%, 26\%)$]{
\label{fig:Head}
\includegraphics[width=0.18\textwidth]{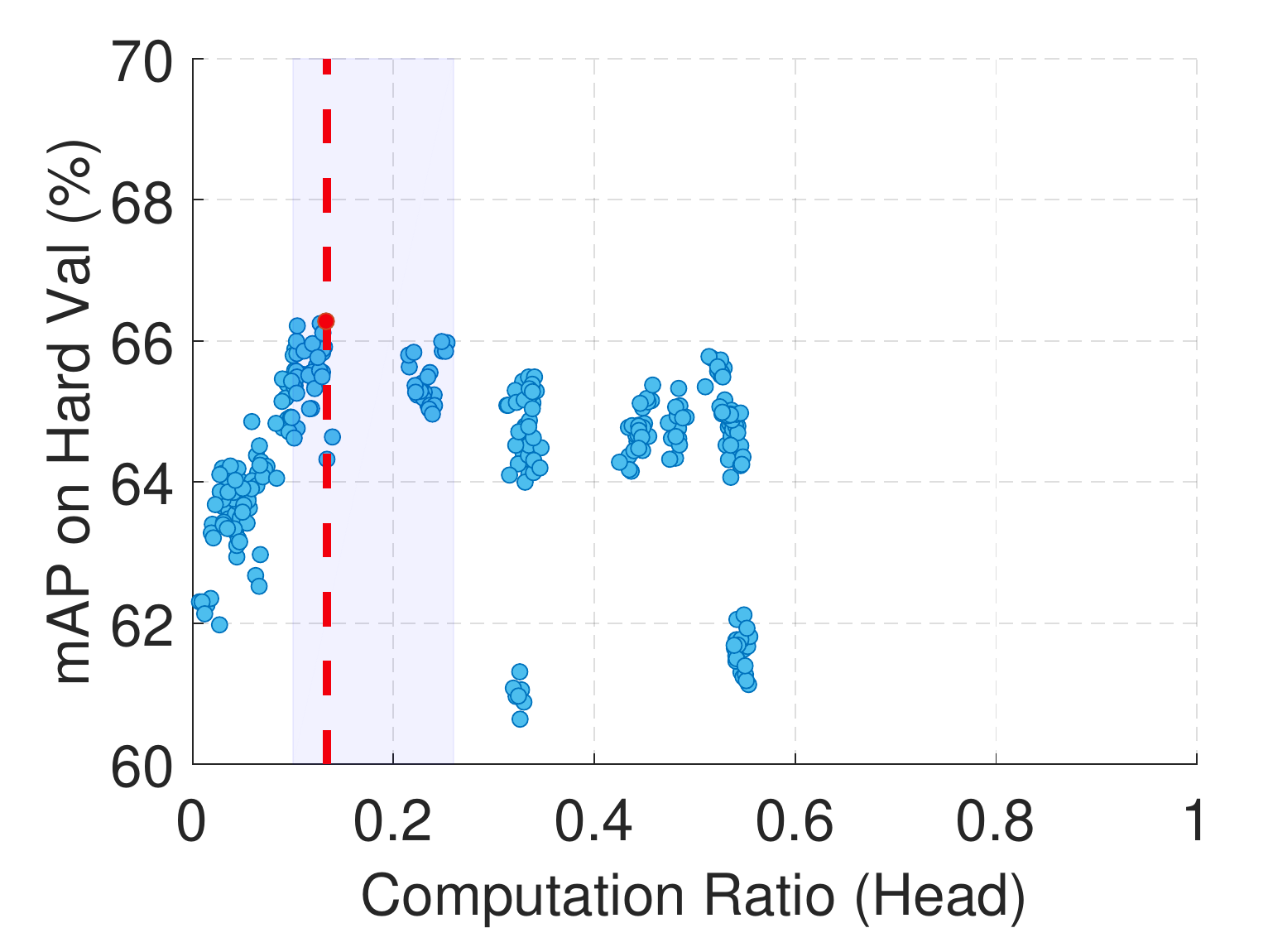}}
\subfigure[Architecture Sketches]{
\label{fig:asketches}
\includegraphics[width=0.4\textwidth]{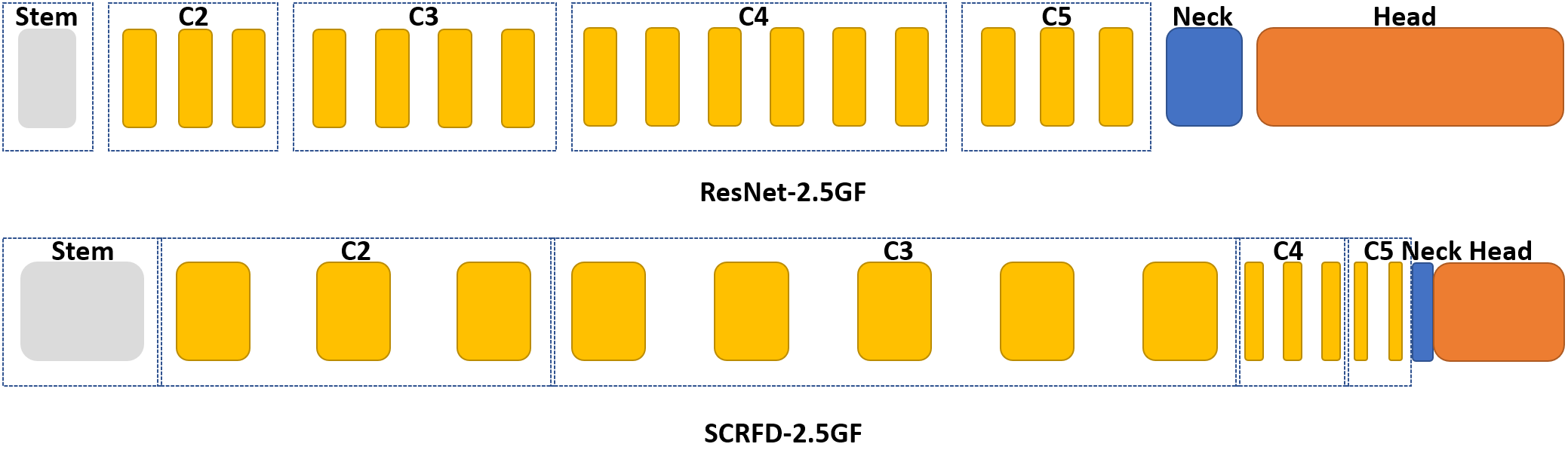}}
\caption{Computation redistribution and architecture sketches under the constraint of 2.5 Gflops. The computation distribution of \scrfdf{2.5} within backbone follows Fig. \ref{fig:computationoptimizationbackbone}. Please refer to Tab. \ref{tab:baselines} for the network configuration of ResNet-2.5GF. In (d), the yellow rectangles in C2 to C5 represents the basic residual block. The width of rectangles corresponds to the computation cost. After computation redistribution, more computations are allocated to shallow stages (\ie C2 and C3).}
\label{fig:computationoptimizationall}
\end{figure*}

\noindent{\bf Computation redistribution on backbone, neck and head.}
After we find the optimised computation distribution within the backbone under a specific computational constraint of 2.5 Gflops, we search for the best computation distribution over the backbone, neck and head. In this step, we only keep the randomly generated network configurations whose backbone settings follow the computation distribution from \scrfdv{1} as shown in Fig. \ref{fig:computationoptimizationbackbone}. 

Now there are another three degrees of freedom (\ie output channel number $n$ for neck, output channel number $h$ for head, and the number of $3\times3$ convolutional layers $m$ in head). We repeat the random sampling in our search space, until we obtain $320$ qualifying models in our target complexity regime (\ie 2.5 Gflops). As evident in Fig. \ref{fig:computationoptimizationall}, most of the computation is allocated in the backbone, with the head following and the neck having the lowest computation ratio. Fig.~\ref{fig:asketches} depicts the comparison of the model architecture, under the constraint of 2.5 Gflops. 
The network configuration of the baseline (ResNet-2.5GF) is introduced in Tab. \ref{tab:baselines}.

By employing the proposed two-step computation redistribution method, we find a large amount of capacity is allocated to the shallow stages, resulting in an AP improvement from $74.47\%$ to $77.87\%$ on the WIDER FACE hard validation set.

\noindent{\bf Higher compute regimes and mobile regime.}
Besides the complexity constraint of 2.5 Gflops, we also utilise the same two-step computation redistribution method to explore the network structure optimisation for higher compute regimes (\eg 10 Gflops and 34 Gflops) and low compute regimes (\eg 0.5 Gflops). In Fig. \ref{fig:resnetscalable} and Fig. \ref{fig:mobilenetscalable}, we show the computation redistribution and the optimised network structures under different computation constraints.

Our final architectures have almost the same flops as the baseline network. From these redistribution results, we can draw the following conclusions: (1) more computation is allocated in the backbone and the computation on the neck and head is compressed; (2) more capacity is reallocated in shallow stages for the 2.5 Gflops, 10 Gflops and 34 Gflops regimes due to the specific scale distribution on WIDER FACE; (3) for the high compute regime (\eg 34 Gflops), the explored structure utilises the bottleneck residual block and we observe significant depth scaling, instead of width scaling in shallow stages. Scaling the width is subject to over-fitting due to the larger increase in parameters \cite{Revisiting2021ResNets}. By contrast, scaling the depth, especially in the earlier layers, introduces fewer parameters compared to scaling the width; (4) for the mobile regime (0.5 Gflops), allocating the limited capacity in the deep stage (\eg C5) for the discriminative features captured in the deep stage, can benefit the shallow small face detection by the top-down neck pathway.

\begin{table}[t!]
\small
\centering
\begin{tabular}{l|cccc}
\hline
Name  & Backbone & Neck & Head  \\
\hline\hline
ResNet-2.5GF        & ResNet34x0.25  & 48   & [96,96]\\
ResNet-10GF         & ResNet34x0.5  & 128  & [160,160]\\
ResNet-34GF         & ResNet50  & 128  & [256,256]\\
\hline
MobileNet-0.5GF     & MobileNetx0.25 & 32   & [80,80]\\
\hline
\end{tabular}
\hspace{1in}
\caption{Network configurations for baselines across different compute regimes. 
$\times 0.25$ and $\times 0.5$ in backbone refer to decreasing the channel number by $0.25$ and $0.5$ compared to the original designs \cite{he2016deep,howard2017mobilenets}. Basic residual blocks are used in ResNet-2.5GF and ResNet-10GF, while bottleneck residual blocks are used in ResNet-34GF. For MobileNet-0.5GF, depth-wise convolution is used in both backbone and head.}
\label{tab:baselines}
\vspace{-4mm}
\end{table}

\begin{table}[t!]
\small
\centering
\begin{tabular}{l|ccc}
\hline
Method  &Easy & Medium & Hard  \\
\hline\hline
ResNet-2.5GF     & 91.87  & 89.49 & 67.32  \\ 
SR + ResNet-2.5GF & 93.21 & 91.11 & 74.47   \\ 
\hline
CR@backbone &  92.32  & 90.25     &  69.78 \\
CR@detector &  92.61  & 90.74     &  70.98 \\
CR@two-step &  92.66  & 90.72     &  71.37 \\
SR+CR@two-step   &  93.78 & 92.16 & 77.87  \\
\hline
\end{tabular}
\hspace{1in}
\caption{Ablation experiments of \scrfdf{2.5} (\ie SR+CR@two-step) on the WIDER FACE validation subset. The baseline model is ResNet-2.5GF as introduced in Tab. \ref{tab:baselines}. ``SR'' and ``CR'' denote the proposed sample and computation redistribution, respectively.}
\label{tab:scrfd25g}
\vspace{-4mm}
\end{table}

\begin{table*}
\begin{center}
\resizebox{1\linewidth}{!}{
\begin{tabular}{cc|ccc|cc|c}
\hline
Method & Backbone  & Easy & Medium & Hard &  \#Params(M) &\#Flops(G) & Infer(ms)\\
\hline\hline
DSFD \cite{li2019dsfd} {\small(CVPR19)}& ResNet152 & 94.29 & 91.47 & 71.39 & 120.06 & 259.55 & 55.6  \\
RetinaFace \cite{deng2019retinaface} {\small(CVPR20)}& ResNet50 & 94.92 & 91.90 & 64.17 & 29.50 & 37.59 & 21.7  \\
HAMBox \cite{liu2019hambox} {\small(CVPR20)}& ResNet50  &  95.27  &  93.76  & 76.75  & 30.24  & 43.28 & 25.9 \\
TinaFace \cite{zhu2020tinaface} {\small(Arxiv20)}& ResNet50 & 95.61 & 94.25 & 81.43 & 37.98 & 172.95 & 38.9  \\
\hline
ResNet-34GF & ResNet50 & 95.64 & 94.22 & 84.02 & 24.81 & 34.16 & 11.8  \\
SCRFD-34GF & Bottleneck Res & 96.06 & 94.92 & 85.29 & 9.80 & 34.13 & 11.7  \\
\hline
ResNet-10GF & ResNet34x0.5 & 94.69 & 92.90 & 80.42 & 6.85 & 10.18 & 6.3  \\ 
SCRFD-10GF & Basic Res & 95.16 & 93.87 & 83.05 & 3.86 & 9.98 & 4.9  \\
\hline
ResNet-2.5GF & ResNet34x0.25 & 93.21 & 91.11 & 74.47 & 1.62 & 2.57 & 5.4  \\ 
SCRFD-2.5GF & Basic Res & 93.78 & 92.16 & 77.87 & 0.67 & 2.53 & 4.2  \\
\hline
\hline
RetinaFace \cite{deng2019retinaface} {\small(CVPR20)}& MobileNet0.25  & 87.78 & 81.16 & 47.32 & 0.44 & 0.802 & 7.9  \\
FaceBoxes \cite{zhang2017faceboxes} (IJCB17) & -  & 76.17 & 57.17 & 24.18 & 1.01 & 0.275 & 2.5  \\
\hline
MobileNet-0.5GF & MobileNetx0.25 & 90.38 & 87.05 & 66.68 & 0.37 & 0.507 & 3.7  \\
SCRFD-0.5GF & Depth-wise Conv & 90.57 & 88.12 & 68.51 & 0.57 & 0.508 & 3.6  \\
\hline
\end{tabular}
}
\end{center}
\caption{Accuracy and efficiency of different methods on the WIDER FACE validation set. The size of test images is bounded by $640$. The proposed sample redistribution is also employed by ResNet-34GF, ResNet-10GF, ResNet-2.5GF and MobileNet-0.5GF for fair comparisons.\#Params and \#Flops denote the number of parameters and multiply-adds. ``Infer'' refers to network inference latency with VGA resolution ($640\times480$) images on NVIDIA 2080TI.}
\vspace{-4mm}
\label{tab:acceff}
\end{table*}

\begin{figure*}[t!]
\centering
\subfigure[Easy]{
\label{fig:valeasy}
\includegraphics[width=0.23\linewidth]{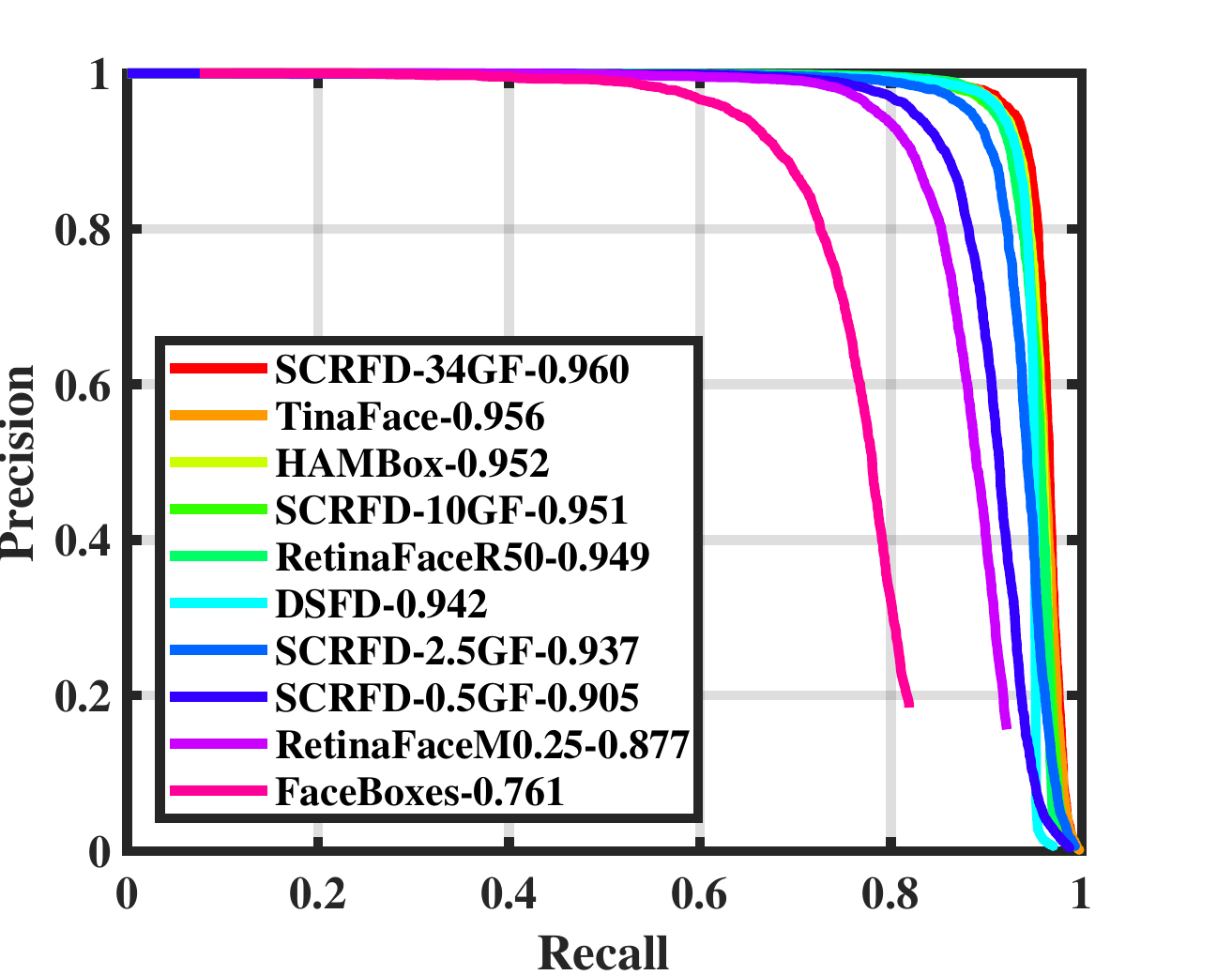}}
\subfigure[Medium]{
\label{fig:valmedium}
\includegraphics[width=0.23\linewidth]{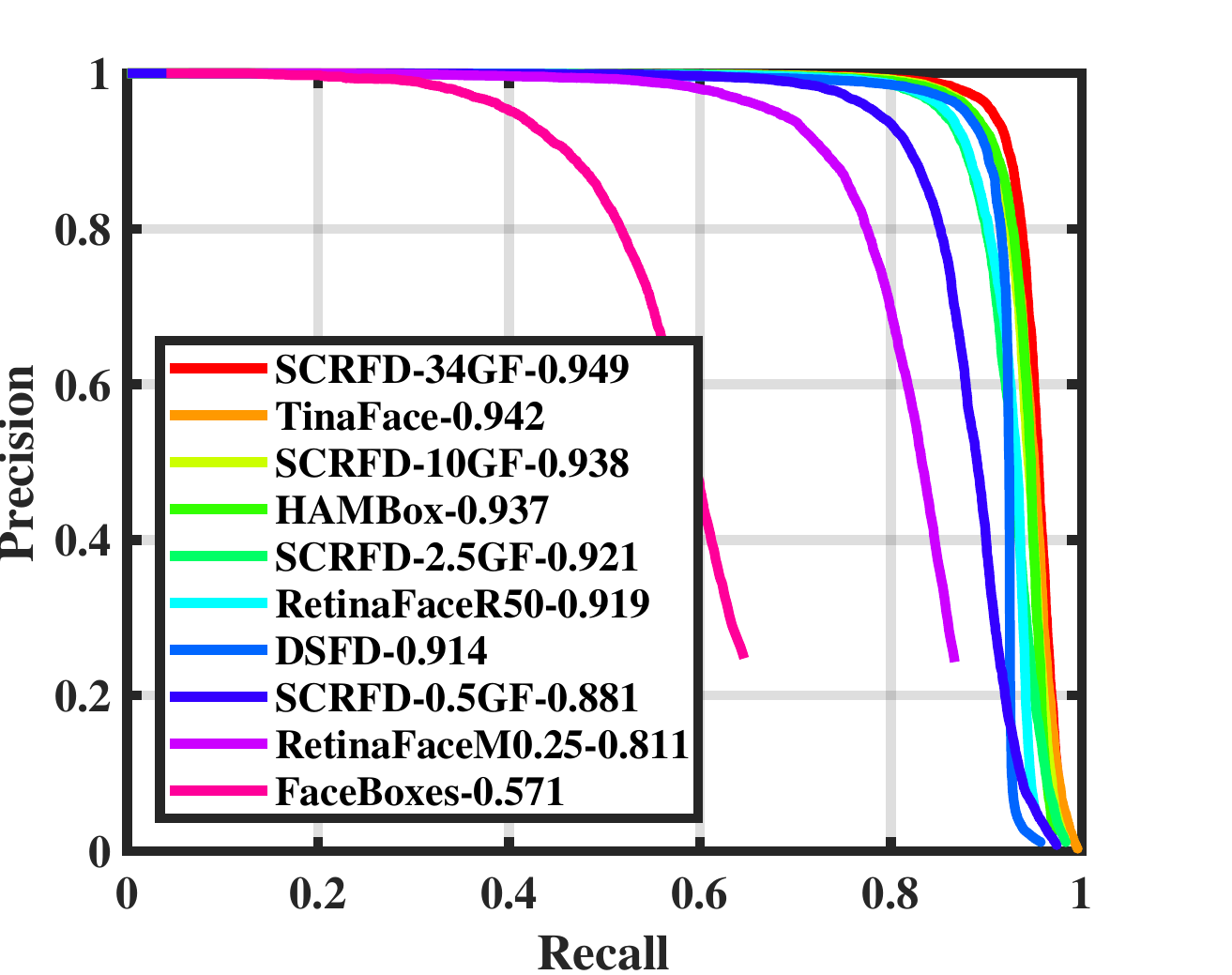}}
\subfigure[Hard]{
\label{fig:valhard}
\includegraphics[width=0.23\linewidth]{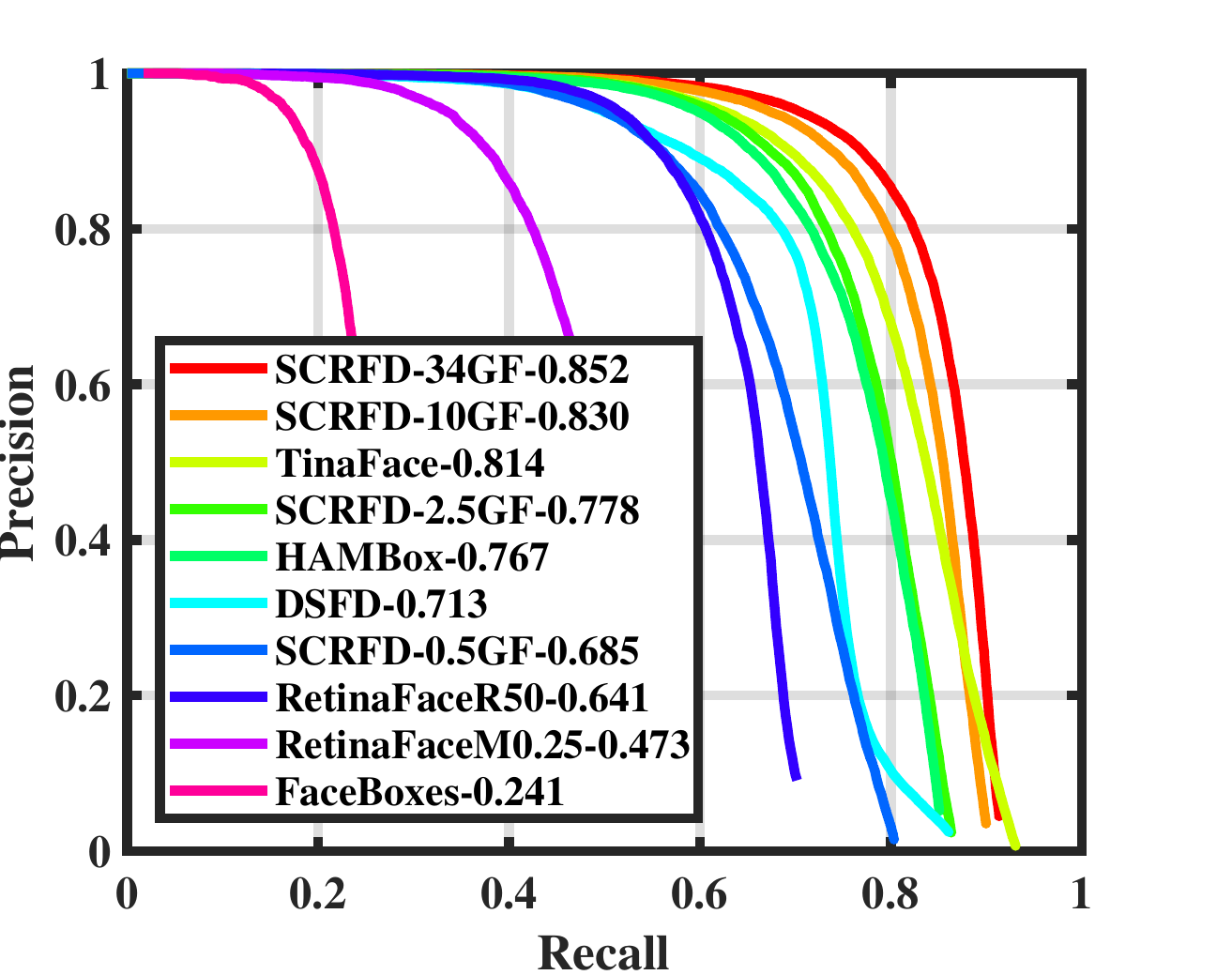}}
\subfigure[Hard@Precision$>98\%$]{
\label{fig:highprecisioncurve}
\includegraphics[width=0.25\linewidth]{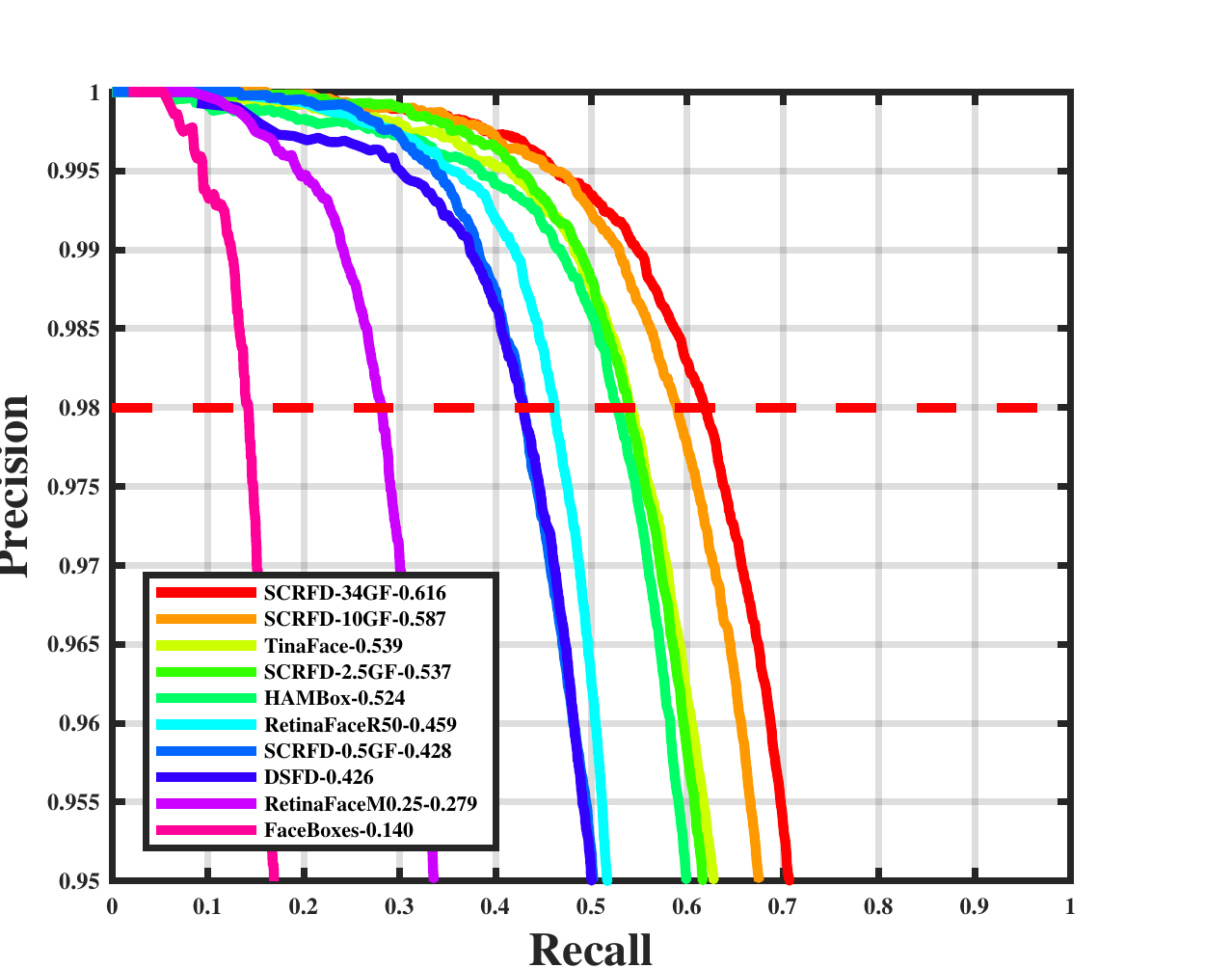}}
\subfigure[Visualisation Results of \scrfdf{2.5}]{
\label{fig:visualisation}
\includegraphics[width=0.9\linewidth]{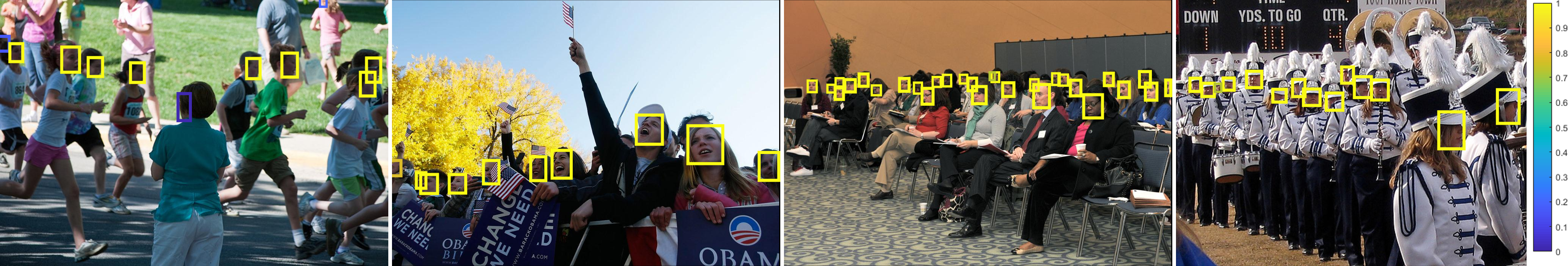}}
\caption{Precision-recall curves and qualitative results on the validation set of the WIDER FACE dataset. Yellow rectangles show the detection results by the proposed \scrfdf{2.5} and brightness encodes the detection confidence.}
\vspace{-4mm}
\label{fig:widerfacerocandvisualisation}
\end{figure*}

\section{Experiments}

\subsection{Implementation Details}

\noindent\textbf{Training.}
For the baseline methods, square patches are cropped from the original images with a random size from $[0.3, 0.45, 0.6, 0.8, 1.0]$ and then these patches are resized to $640\times640$ for training. For the proposed Sample Redistribution (SR), we enlarge the random size set by adding $[1.2, 1.4, 1.6, 1.8, 2.0]$ scales. Besides scale augmentation, the training data are also augmented by color distortion and random horizontal flipping, with a probability of $0.5$. For the anchor setting, we tile anchors of $\{16, 32\}$, $\{64, 128\}$, and $\{256, 512\}$ on the feature maps of stride 8, 16, and 32, respectively. The anchor ratio is set as $1.0$. In this paper, we employ Adaptive Training Sample Selection (ATSS) \cite{zhang2020bridging} for positive anchor matching. In the detection head, weight sharing and Group Normalisation \cite{wu2018group} are used \cite{tian2019fcos}.
The losses of classification and regression branches are Generalised Focal Loss (GFL) \cite{li2020generalized} and DIoU loss \cite{zheng2020distance}, respectively. 

Our experiments are implemented in PyTorch, based on the open-source mmdetection \cite{chen2019mmdetection}. 
We adopt the SGD optimizer (momentum 0.9, weight decay 5e-4) with a batch size of $8\times 4$ and train on four Tesla V100. 
The initial learning rate is set to $0.00001$, linearly warming up to $0.01$ within the first $3$ epochs. During network search, the learning rate is multiplied by $0.1$ at the $56$-th, and $68$-th epochs. The learning process terminates on the $80$-th epoch. For re-training of both baselines and the best configurations after search, the learning rate decays by a factor of 10 at the $440$-th and $544$-th epochs, and the learning process terminates at the $640$-th epoch. For the computation constrained baselines in Tab.~\ref{tab:baselines}, we also employ the same training schedule. All the models are trained from scratch without any pre-training.

\noindent\textbf{Testing.}
We only employ single-scale testing, bounded by the VGA resolution ($640\times480$). 
The results of DSFD \cite{li2019dsfd}, RetinaFace \cite{deng2019retinaface}, TinaFace \cite{zhu2020tinaface} and Faceboxes \cite{zhang2017faceboxes} are reported by testing the released models, while the HAMBox \cite{liu2019hambox} model is shared from the author.

\subsection{Ablation Study}
In Tab. \ref{tab:scrfd25g} we present the AP performance of models by gradually including the proposed sample and computation redistribution methods. Our baseline model (\ie ResNet-2.5GF as introduced in Tab.~\ref{tab:baselines}) gets $91.87\%$, $89.49\%$, and $67.32\%$ in the three settings on the validation subset. By adding sample redistribution, the hard set AP significantly increases to $74.47\%$, indicating the benefit from allocating more training samples on the feature map of stride 8. 

After separately employing the proposed computation redistribution on the backbone and the whole detector, the AP on the hard set improves to $69.78\%$ and $70.98\%$. This indicates that (1) the network structure directly inherited from the classification task is sub-optimal for the face detection task, and (2) joint computation reallocation on the backbone, neck and head outperforms computation optimisation applied only on the backbone. Furthermore, the proposed two-step computation redistribution strategy achieves AP of $71.37\%$, surpassing one-step computation reallocation on the whole detector by $0.39\%$. As we shrink the whole search space by the proposed two-step strategy and our random model sampling number is fixed at $320$, the two-step method is possible to find better network configurations from the large search space. Since sample redistribution and computation redistribution are orthogonal, their combination (\ie \scrfdf{2.5}) shows a further improvement, achieving AP of $77.87\%$ on the hard track.

\subsection{Accuracy and Efficiency on WIDER FACE}
As shown in Tab.~\ref{tab:acceff} and Fig.~\ref{fig:widerfacerocandvisualisation}, we compared the proposed method with other state-of-the-art face detection algorithms (\eg DSFD \cite{li2019dsfd}, RetinaFace \cite{deng2019retinaface}, HAMBox \cite{liu2019hambox} and TinaFace \cite{zhu2020tinaface}). Since we fix the testing scale at $640$, the flops and inference time directly correspond to the reported AP. The proposed \scrfdf{34} outperforms all these state-of-the-art methods on the three subsets, especially for the hard track, which contains a large number of tiny faces. More specifically, \scrfdf{34} surpasses TinaFace by $3.86\%$ while being more than \emph{3$\times$ faster} on GPUs. In addition, the computation cost of \scrfdf{34} is only around $20\%$ of TinaFace. As \scrfdf{34} scales the depth in the earlier layers, it also introduces fewer parameters, resulting in a much smaller model size ($9.80M$). 

Overall, all of the proposed \scrfd models (\eg \scrfdf{34}, \scrfdf{10}, \scrfdf{2.5} and \scrfdf{0.5}) provide considerable improvements compared to the baseline models (listed in Tab. \ref{tab:baselines}), by optimising the network structure alone (\ie computation redistribution), across a wide range of compute regimes. For the low-compute regimes, \scrfdf{0.5} significantly outperforms RetinaFaceM0.25 by $21.19\%$ on the hard AP, while consuming only $63.34\%$ computation and $45.57\%$ inference time. 

For real-world face detection system, high precision (\eg $>98\%$) is required to avoid frequent false alarms. As shown in Fig. \ref{fig:highprecisioncurve},
\scrfdf{2.5} obtains comparable AP ($53.7\%$) compared to TinaFace ($53.9\%$) when the threshold score is improved to achieve prevision higher than $98\%$, while the computation cost is only $1.46\%$ and the inference time is only $10.8\%$. Fig.~\ref{fig:visualisation} shows qualitative results generated by \scrfdf{2.5}. As can be seen, our face detector works very well in both indoor and outdoor crowded scenes under different conditions (\eg appearance variations from pose, occlusion and illumination). The impressive performance across a wide range of scales indicate that \scrfdf{2.5} has a very high recall and can detect faces accurately even without large scale testing.

\section{Conclusions}

In this work, we present a sample and computation redistribution paradigm for efficient face detection. Our results show significantly improved accuracy and efficiency trade-off by the proposed \scrfd across a wide range of compute regimes, when compared to the current state-of-the-art.

{\small
\bibliographystyle{ieee_fullname}
\bibliography{egbib}

\begin{thebibliography}{10}\itemsep=-1pt

\bibitem{Revisiting2021ResNets}
Irwan Bello, William Fedus, Xianzhi Du, Ekin Cubuk, Aravind Srinivas, Tsung-Yi
  Lin, Jonathon Shlens, and Barret Zoph.
\newblock Revisiting resnets: Improved training and scaling strategies.
\newblock {\em arXiv:2103.07579}, 2021.

\bibitem{bulat2017far}
Adrian Bulat and Georgios Tzimiropoulos.
\newblock How far are we from solving the 2d \& 3d face alignment problem?(and
  a dataset of 230,000 3d facial landmarks).
\newblock In {\em ICCV}, 2017.

\bibitem{chen2019mmdetection}
Kai Chen, Jiaqi Wang, Jiangmiao Pang, Yuhang Cao, Yu Xiong, Xiaoxiao Li,
  Shuyang Sun, Wansen Feng, Ziwei Liu, Jiarui Xu, et~al.
\newblock Mmdetection: Open mmlab detection toolbox and benchmark.
\newblock {\em arXiv:1906.07155}, 2019.

\bibitem{chen2019detnas}
Yukang Chen, Tong Yang, Xiangyu Zhang, Gaofeng Meng, Chunhong Pan, and Jian
  Sun.
\newblock Detnas: Neural architecture search on object detection.
\newblock {\em arXiv:1903.10979}, 2019.

\bibitem{deng2009imagenet}
Jia Deng, Wei Dong, Richard Socher, Li-Jia Li, Kai Li, and Li Fei-Fei.
\newblock Imagenet: A large-scale hierarchical image database.
\newblock In {\em CVPR}, 2009.

\bibitem{deng2018arcface}
Jiankang Deng, Jia Guo, Niannan Xue, and Stefanos Zafeiriou.
\newblock Arcface: Additive angular margin loss for deep face recognition.
\newblock In {\em CVPR}, 2019.

\bibitem{deng2019retinaface}
Jiankang Deng, Jia Guo, Yuxiang Zhou, Jinke Yu, Irene Kotsia, and Stefanos
  Zafeiriou.
\newblock Retinaface: Single-stage dense face localisation in the wild.
\newblock In {\em CVPR}, 2020.

\bibitem{Efron1994}
Bradley Efron and Robert~J Tibshirani.
\newblock {\em An introduction to the bootstrap}.
\newblock CRC press, 1994.

\bibitem{feng2018joint}
Yao Feng, Fan Wu, Xiaohu Shao, Yanfeng Wang, and Xi Zhou.
\newblock Joint 3d face reconstruction and dense alignment with position map
  regression network.
\newblock In {\em ECCV}, 2018.

\bibitem{ghiasi2019fpn}
Golnaz Ghiasi, Tsung-Yi Lin, and Quoc~V Le.
\newblock Nas-fpn: Learning scalable feature pyramid architecture for object
  detection.
\newblock In {\em CVPR}, 2019.

\bibitem{guo2019single}
Zichao Guo, Xiangyu Zhang, Haoyuan Mu, Wen Heng, Zechun Liu, Yichen Wei, and
  Jian Sun.
\newblock Single path one-shot neural architecture search with uniform
  sampling.
\newblock {\em arXiv:1904.00420}, 2019.

\bibitem{he2016deep}
Kaiming He, Xiangyu Zhang, Shaoqing Ren, and Jian Sun.
\newblock Deep residual learning for image recognition.
\newblock In {\em CVPR}, 2016.

\bibitem{he2019bag}
Tong He, Zhi Zhang, Hang Zhang, Zhongyue Zhang, Junyuan Xie, and Mu Li.
\newblock Bag of tricks for image classification with convolutional neural
  networks.
\newblock In {\em CVPR}, 2019.

\bibitem{howard2017mobilenets}
Andrew~G Howard, Menglong Zhu, Bo Chen, Dmitry Kalenichenko, Weijun Wang,
  Tobias Weyand, Marco Andreetto, and Hartwig Adam.
\newblock Mobilenets: Efficient convolutional neural networks for mobile vision
  applications.
\newblock {\em arXiv:1704.04861}, 2017.

\bibitem{li2019dsfd}
Jian Li, Yabiao Wang, Changan Wang, Ying Tai, Jianjun Qian, Jian Yang, Chengjie
  Wang, Jilin Li, and Feiyue Huang.
\newblock Dsfd: dual shot face detector.
\newblock In {\em CVPR}, 2019.

\bibitem{li2020generalized}
Xiang Li, Wenhai Wang, Lijun Wu, Shuo Chen, Xiaolin Hu, Jun Li, Jinhui Tang,
  and Jian Yang.
\newblock Generalized focal loss: Learning qualified and distributed bounding
  boxes for dense object detection.
\newblock {\em arXiv:2006.04388}, 2020.

\bibitem{liang2019computation}
Feng Liang, Chen Lin, Ronghao Guo, Ming Sun, Wei Wu, Junjie Yan, and Wanli
  Ouyang.
\newblock Computation reallocation for object detection.
\newblock In {\em ICLR}, 2020.

\bibitem{lin2017feature}
Tsung-Yi Lin, Piotr Doll{\'a}r, Ross Girshick, Kaiming He, Bharath Hariharan,
  and Serge Belongie.
\newblock Feature pyramid networks for object detection.
\newblock In {\em CVPR}, 2017.

\bibitem{lin2017focal}
Tsung-Yi Lin, Priya Goyal, Ross Girshick, Kaiming He, and Piotr Doll{\'a}r.
\newblock Focal loss for dense object detection.
\newblock In {\em ICCV}, 2017.

\bibitem{lin2014microsoft}
Tsung-Yi Lin, Michael Maire, Serge Belongie, James Hays, Pietro Perona, Deva
  Ramanan, Piotr Doll{\'a}r, and C~Lawrence Zitnick.
\newblock Microsoft coco: Common objects in context.
\newblock In {\em ECCV}, 2014.

\bibitem{liu2018path}
Shu Liu, Lu Qi, Haifang Qin, Jianping Shi, and Jiaya Jia.
\newblock Path aggregation network for instance segmentation.
\newblock In {\em CVPR}, 2018.

\bibitem{liu2020bfbox}
Yang Liu and Xu Tang.
\newblock Bfbox: Searching face-appropriate backbone and feature pyramid
  network for face detector.
\newblock In {\em CVPR}, 2020.

\bibitem{liu2019hambox}
Yang Liu, Xu Tang, Xiang Wu, Junyu Han, Jingtuo Liu, and Errui Ding.
\newblock Hambox: Delving into online high-quality anchors mining for detecting
  outer faces.
\newblock {\em CVPR}, 2020.

\bibitem{ming2019group}
Xiang Ming, Fangyun Wei, Ting Zhang, Dong Chen, and Fang Wen.
\newblock Group sampling for scale invariant face detection.
\newblock In {\em CVPR}, 2019.

\bibitem{najibi2017ssh}
Mahyar Najibi, Pouya Samangouei, Rama Chellappa, and Larry~S Davis.
\newblock Ssh: Single stage headless face detector.
\newblock In {\em ICCV}, 2017.

\bibitem{pan2018mean}
Hongyu Pan, Hu Han, Shiguang Shan, and Xilin Chen.
\newblock Mean-variance loss for deep age estimation from a face.
\newblock In {\em CVPR}, 2018.

\bibitem{radosavovic2020designing}
Ilija Radosavovic, Raj~Prateek Kosaraju, Ross Girshick, Kaiming He, and Piotr
  Doll{\'a}r.
\newblock Designing network design spaces.
\newblock In {\em CVPR}, 2020.

\bibitem{schroff2015facenet}
Florian Schroff, Dmitry Kalenichenko, and James Philbin.
\newblock Facenet: A unified embedding for face recognition and clustering.
\newblock In {\em CVPR}, 2015.

\bibitem{szegedy2015going}
Christian Szegedy, Wei Liu, Yangqing Jia, Pierre Sermanet, Scott Reed, Dragomir
  Anguelov, Dumitru Erhan, Vincent Vanhoucke, and Andrew Rabinovich.
\newblock Going deeper with convolutions.
\newblock In {\em CVPR}, 2015.

\bibitem{tang2018pyramidbox}
Xu Tang, Daniel~K Du, Zeqiang He, and Jingtuo Liu.
\newblock Pyramidbox: A context-assisted single shot face detector.
\newblock In {\em ECCV}, 2018.

\bibitem{tian2019fcos}
Zhi Tian, Chunhua Shen, Hao Chen, and Tong He.
\newblock Fcos: Fully convolutional one-stage object detection.
\newblock In {\em ICCV}, 2019.

\bibitem{viola2004robust}
Paul Viola and Michael~J Jones.
\newblock Robust real-time face detection.
\newblock {\em IJCV}, 2004.

\bibitem{wu2018group}
Yuxin Wu and Kaiming He.
\newblock Group normalization.
\newblock In {\em ECCV}, 2018.

\bibitem{yang2016wider}
Shuo Yang, Ping Luo, Chen-Change Loy, and Xiaoou Tang.
\newblock Wider face: A face detection benchmark.
\newblock In {\em CVPR}, 2016.

\bibitem{zhang2020asfd}
Bin Zhang, Jian Li, Yabiao Wang, Ying Tai, Chengjie Wang, Jilin Li, Feiyue
  Huang, Yili Xia, Wenjiang Pei, and Rongrong Ji.
\newblock Asfd: Automatic and scalable face detector.
\newblock {\em arXiv:2003.11228}, 2020.

\bibitem{zhang2018jointexpression}
Feifei Zhang, Tianzhu Zhang, Qirong Mao, and Changsheng Xu.
\newblock Joint pose and expression modeling for facial expression recognition.
\newblock In {\em CVPR}, 2018.

\bibitem{zhang2020bridging}
Shifeng Zhang, Cheng Chi, Yongqiang Yao, Zhen Lei, and Stan~Z Li.
\newblock Bridging the gap between anchor-based and anchor-free detection via
  adaptive training sample selection.
\newblock In {\em CVPR}, 2020.

\bibitem{zhang2017faceboxes}
Shifeng Zhang, Xiangyu Zhu, Zhen Lei, Hailin Shi, Xiaobo Wang, and Stan~Z Li.
\newblock Faceboxes: A cpu real-time face detector with high accuracy.
\newblock In {\em IJCB}, 2017.

\bibitem{zhang2017s3fd}
Shifeng Zhang, Xiangyu Zhu, Zhen Lei, Hailin Shi, Xiaobo Wang, and Stan~Z Li.
\newblock S3fd: Single shot scale-invariant face detector.
\newblock In {\em ICCV}, 2017.

\bibitem{zheng2020distance}
Zhaohui Zheng, Ping Wang, Wei Liu, Jinze Li, Rongguang Ye, and Dongwei Ren.
\newblock Distance-iou loss: Faster and better learning for bounding box
  regression.
\newblock In {\em AAAI}, 2020.

\bibitem{zhu2020tinaface}
Yanjia Zhu, Hongxiang Cai, Shuhan Zhang, Chenhao Wang, and Yichao Xiong.
\newblock Tinaface: Strong but simple baseline for face detection.
\newblock {\em arXiv:2011.13183}, 2020.

\end{thebibliography}
}

\end{document}